\documentclass[letterpaper, 10 pt, journal, twoside]{IEEEtran}

\usepackage{times}
\usepackage{blindtext}
\usepackage{amsmath,amssymb,amsfonts}
\usepackage{graphicx}
\usepackage{units}
\usepackage{algorithm}
\usepackage[noend]{algpseudocode}
\usepackage{dblfloatfix}    
\usepackage{graphicx, xfrac}
\usepackage{textcomp}
\usepackage{xcolor}
\makeatletter
\let\NAT@parse\undefined
\makeatother
\usepackage[colorlinks]{hyperref}
\hypersetup{citecolor = blue, linkcolor = blue, urlcolor=black}
\usepackage[T1]{fontenc}
\usepackage{orcidlink}
\usepackage{parnotes}
\usepackage[caption=false,font=footnotesize,labelfont=sf,textfont=sf]{subfig}
\usepackage[flushleft]{threeparttable}
\graphicspath{{./figures/}}

\usepackage{array}

\begin{document}
%
\title{All-UWB SLAM Using UWB Radar and \\ UWB AOA}
%
%
%


\author{Charith Premachandra\raisebox{0.5ex}{\orcidlink{0000-0002-8662-4652}}, 
	Achala Athukorala\raisebox{0.5ex}{\orcidlink{0000-0003-1908-4980}}, \IEEEmembership{Member,~IEEE,} and U-Xuan Tan\raisebox{0.5ex}{\orcidlink{0000-0002-5757-1379}}, \IEEEmembership{Senior Member,~IEEE}%
\thanks{Manuscript received 16 February 2025; Revised 15 May 2025; Accepted 3 June 2025.}
\thanks{This paper was recommended for publication by Editor J. Civera upon evaluation of the Associate Editor and Reviewers' comments.} 

\thanks{The authors are with the Singapore
	University of Technology and Design, Singapore 485998 (e-mail:
	gihan\_appuhamilage@mymail.sutd.edu.sg; achala\_chathuranga@sutd.edu.sg; uxuan\_tan@sutd.edu.sg).} %
\thanks{Digital Object Identifier (DOI): see top of this page.}
}

%
%

\markboth{IEEE Robotics and Automation Letters. Preprint Version. Accepted June, 2025}
{Premachandra \MakeLowercase{\textit{et al.}}: All-UWB SLAM Using UWB Radar and UWB AOA} 

%



\maketitle

\begin{abstract}
There has been a growing interest in autonomous systems designed to operate in adverse conditions (e.g. smoke, dust), where the visible light spectrum fails.
In this context, Ultra-wideband (UWB) radar is capable of penetrating through such challenging environmental conditions due to the lower frequency components within its broad bandwidth. 
Therefore, UWB radar has emerged as a potential sensing technology for Simultaneous Localization and Mapping (SLAM) in vision-denied environments where optical sensors (e.g. LiDAR, Camera) are prone to failure.
Existing approaches involving UWB radar as the primary exteroceptive sensor generally extract features in the environment, which are later initialized as landmarks in a map. 
However, these methods are constrained by the number of distinguishable features in the environment. Hence, this paper proposes a novel method incorporating UWB Angle of Arrival (AOA) measurements into UWB radar-based SLAM systems to improve the accuracy and scalability of SLAM in feature-deficient environments. 
The AOA measurements are obtained using UWB anchor-tag units which are dynamically deployed by the robot in featureless areas during mapping of the environment.
This paper thoroughly discusses prevailing constraints associated with UWB
AOA measurement units and presents solutions to overcome them.
Our experimental results show that integrating UWB AOA  units with UWB radar enables SLAM in vision-denied feature-deficient environments.

\end{abstract}


%
\IEEEpeerreviewmaketitle

\section{Introduction}
%
%
%
%
\IEEEPARstart{S}{imultaneous} Localization and Mapping (SLAM) is a crucial component that enables mobile robots to navigate autonomously in unknown environments. 
\textcolor{black}{SLAM and navigation systems often use optical sensors such as LiDAR or camera as the primary exteroceptive sensor \cite{RN378, 10320444}.} 
These optical sensors operate near the visible light region of the electromagnetic spectrum, and are thus susceptible to failure in adverse environmental conditions (e.g. fog, dust, rain). 
On the other hand, Radio Frequency (RF) signals are capable of penetrating such conditions due to their longer wavelengths. 
As a result, radar technologies such as Frequency Modulated Continuous Waves (FMCW) and Ultra-wideband (UWB) radar are being explored as alternatives for perception sensors, especially in vision-denied scenarios \cite{RN339, smoke, cp1, liu2024rangeslamultrawidebandbasedsmokeresistantrealtime}.

UWB radar is preferred over FMCW radar due to its high signal-to-noise ratio (SNR) \cite{RN83}.
Unlike FMCW radar, state-of-the-art UWB radar systems output the received signal strength as a time-series instead of a point cloud. The time-series data are further processed to extract features in the environment \cite{cp1}, or to perform fingerprinting-based SLAM \cite{RN60}.
Landmark-based UWB radar SLAM is preferred over fingerprinting approaches due to better representation of the environment and lower computational load.  Generally, an array of UWB radar sensors is used to extract static features in the environment which are later initialized as landmarks at the SLAM backend \cite{cp1, RN123, RN139}. 

UWB anchor-tag systems are generally used in measuring the relative distance between nodes using the time-of-flight (TOF) principle.
Such anchor-tag ranging systems have been used in localization frameworks, where three or more anchors are fixed in the environment to localize the tags \cite{radio}.
Recently, single anchor systems have also been introduced for localization of tags in the environment \cite{feng}. 
These anchor-tag systems are fused with common existing exteroceptive sensors such as LiDAR \cite{zhou} and camera \cite{RN58} to improve the performance of SLAM systems in mobile robots. 
The exteroceptive sensors localize the robot relative to the local environment, while the UWB-based localization helps correct the accumulated pose estimation error in the long run, thereby improving localization and mapping accuracy.

\begin{figure}[!t]
	\centering
	{\includegraphics[height=1.75in]{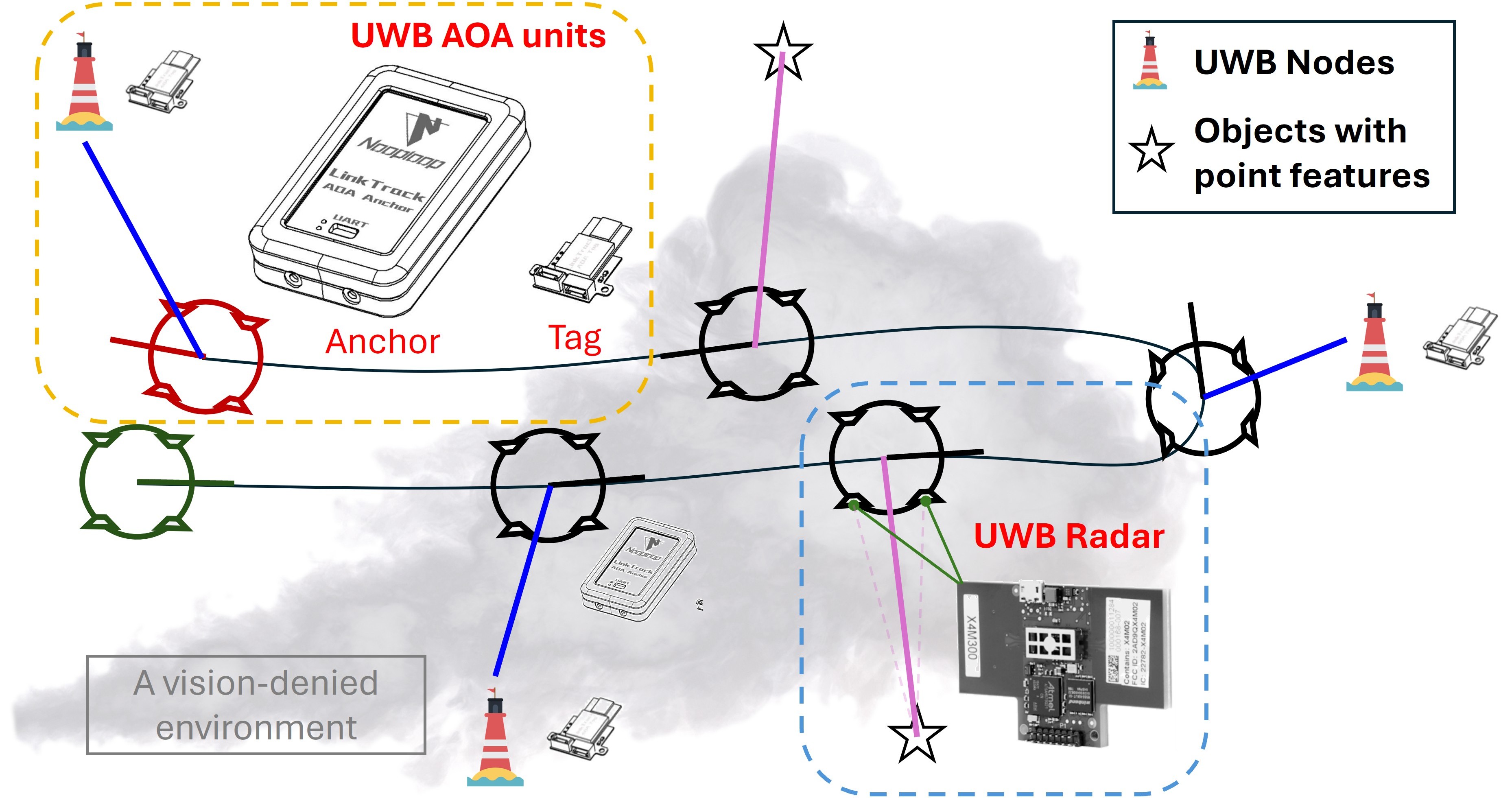}}
	\caption{\textbf{All-UWB SLAM}: The proposed system comprises UWB radar as the primary exteroceptive sensor to perceive the surroundings in vision-denied environments. Meanwhile, UWB AOA nodes are deployed by the mobile robot when needed in a feature-deficient area.}
	\label{abstract}
\end{figure}

\begin{figure}[!t]
	\centering
	{\includegraphics[height=1.82in]{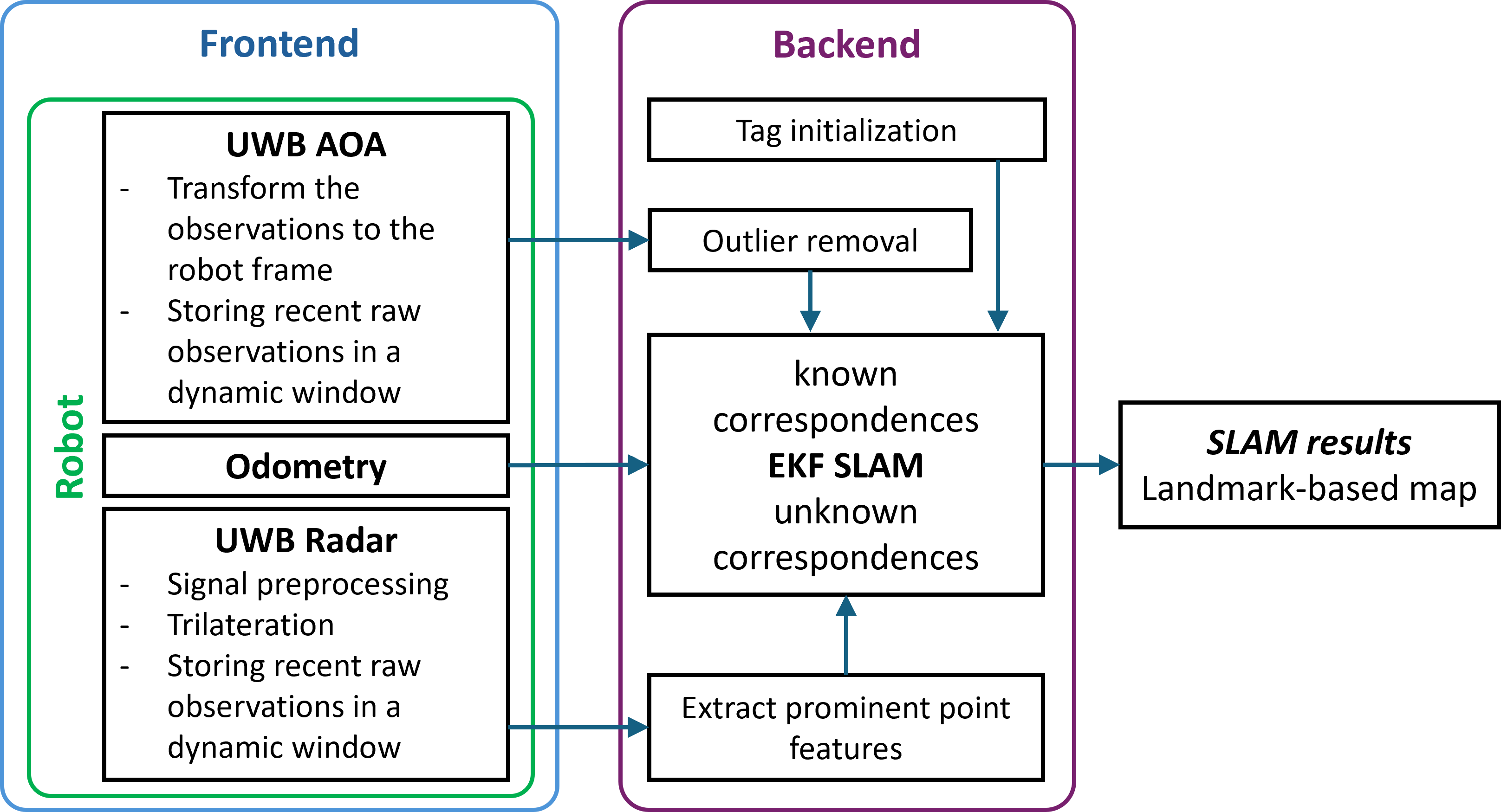}}
	\caption{\textbf{Overview of the proposed SLAM system.} There are two main sensors at the frontend: UWB AOA anchor-tag modules and UWB radars.
		The backend generates a landmark-based map that comprises deployed tags and prominent point features in the environment.}
	\label{oview}
\end{figure}

\begin{figure}[!t]
	\centering
	{\includegraphics[clip, trim={1.2cm 0 0 0}, width=3.4in]{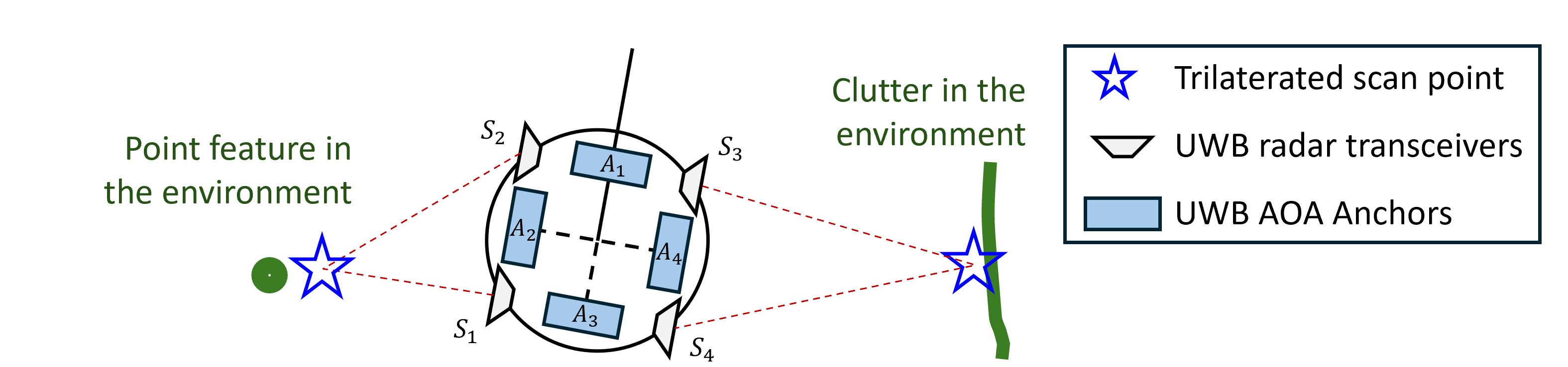}}
	\caption{\textbf{Plan view of the proposed sensor configuration.} Two arrays of UWB Radar modules are installed on both sides of a non-holonomic mobile robot. Another array of UWB AOA anchors are installed on top of the robot covering the entire 360$^\circ$ view.}
	\label{sensor_config}
\end{figure}

Compared to TOF-only systems, angle-of-arrival (AOA) measurements provide both relative angle and the distance between two nodes \cite{RN379}. Hence, by employing bearing observations alongside range measurements, an AOA system is expected to outperform radio SLAM systems that rely solely on range or bearing data.
However, UWB AOA systems are yet to be established in the SLAM domain. Although several studies have utilized UWB AOA anchor-tags, their practicality in real-world applications remains uncertain due to the inherent noise in observations caused by multi-path propagation and non-line-of-sight (NLOS) conditions.

Considering these facts, this paper proposes a novel UWB-based SLAM approach which fuses UWB AOA anchor-tags with UWB radar-based SLAM for vision-denied environments as shown in Fig. \ref{abstract}.
The main contributions of this paper are summarized as follows:
\begin{enumerate}
	\item We propose a SLAM system which incorporates onboard UWB radar, and dynamic deployment of UWB AOA tag modules to perform SLAM in vision-denied feature-deficient environments;
	\item We propose a moving window-based approach to extract only the prominent point features in the environment using UWB radar. The same window is used for filtering multi-path propagation in UWB AOA readings;
	\item We implement the proposed SLAM system in ROS2, and achieved real-time performance. The experimental results demonstrate the performance of our proposed SLAM system in feature-deficient environments.
	
\end{enumerate}

The paper is organized as follows. 
In Section \ref{sec::rel_work}, we review existing work on UWB-based SLAM systems, followed by our proposed approach in Section \ref{sec::prop_appr}.
We present real-world experimental results in Section \ref{sec::exp_and_res}, and conclude the paper outlining future directions in Section \ref{sec::conclusions}.

\section{Related Work} \label{sec::rel_work}

Recently, UWB radar has increasingly been used in challenging environments (e.g. dense smoke) for fire rescue \cite{fire}, floor plan construction \cite{smoke}, and SLAM \cite{cp1}.   
However, existing UWB radar-based SLAM systems assume that the environment contains sufficient number of distinguishable features, and have only been demonstrated in structured simple environments with dense features \cite{cp1, RN123, RN139}. 
Insufficient features pose challenges for loop closure, creating duplicate landmarks in previously explored areas.
This issue is particularly problematic in cluttered, feature-deficient real-world environments, where reliable data association becomes difficult.
Clutter compromises existing UWB radar-based SLAM systems due to multi-path propagation and less distinguishable features. 

\begin{figure}[!t]
	\centering
	{\includegraphics[height=1.1in]{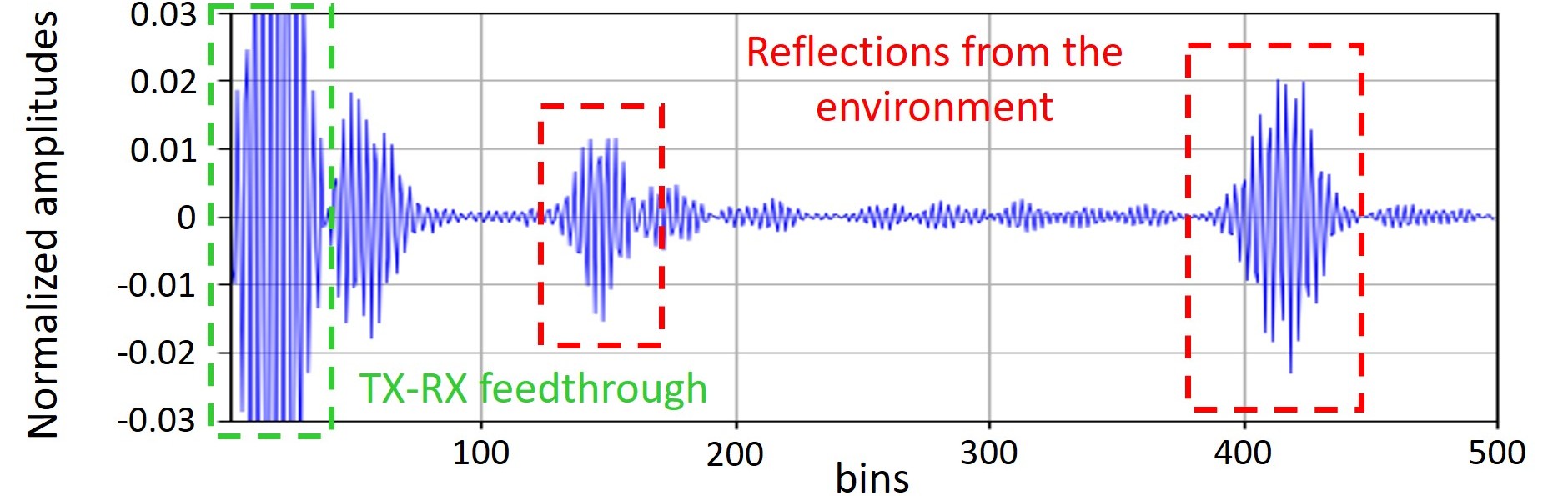}}
	\caption{\textbf{Raw UWB radar observation}: The output of Novelda's X4M300 UWB radar is a time series. One sample size (i.e. 1 bin) corresponds to 6.4 mm after converting the time domain to the distance domain using the speed of light considering TOF. High amplitudes correspond to the reflections from the environment from objects in the radar's field of view (FOV). 
	}
	\label{raw_signal}
\end{figure}

\begin{figure*}[!t]
	\centering
	\subfloat[]{\includegraphics[height=1.75in]{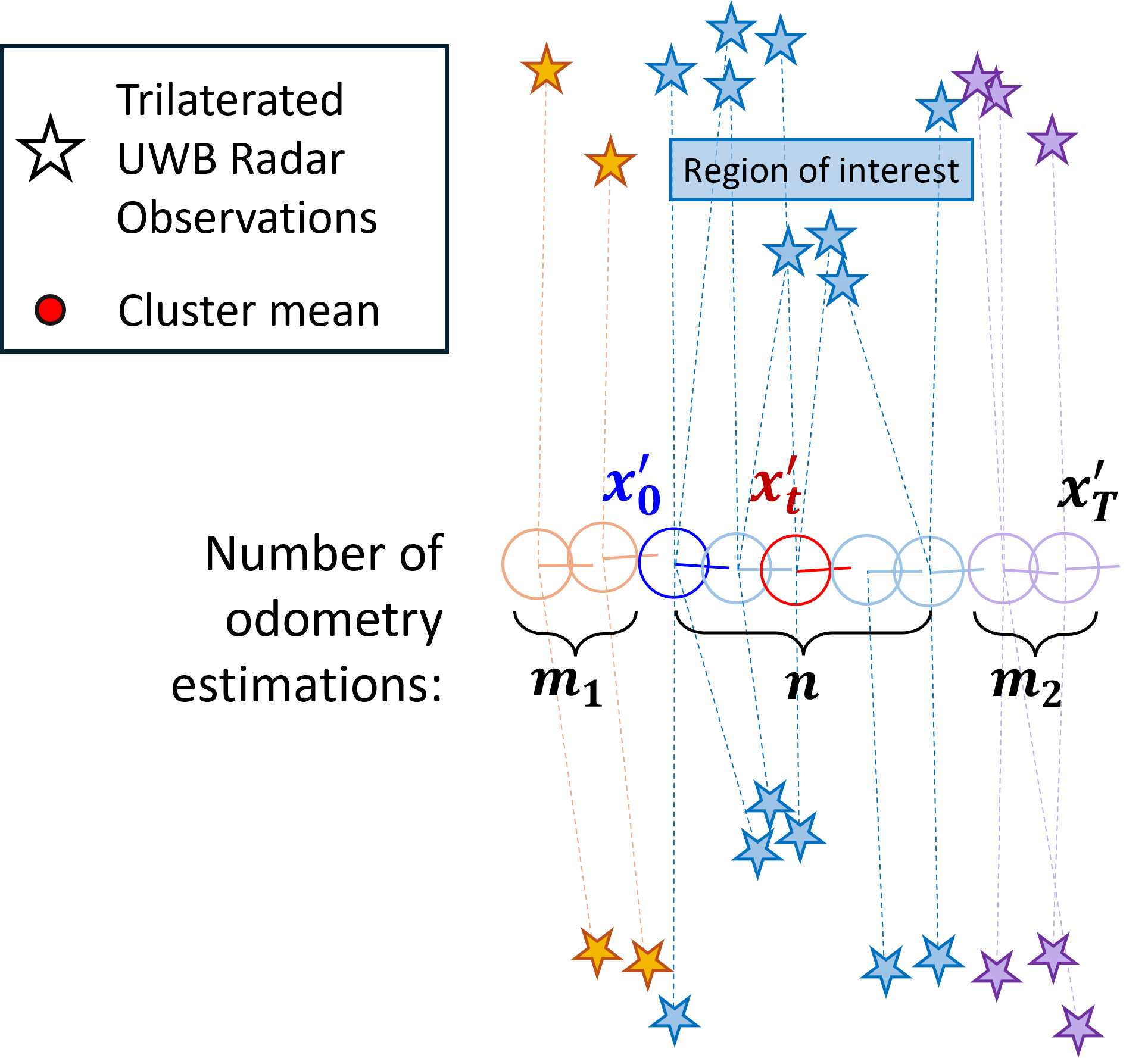}%
		\label{step1}}
	\hfil
	\subfloat[]{\includegraphics[height=1.75in]{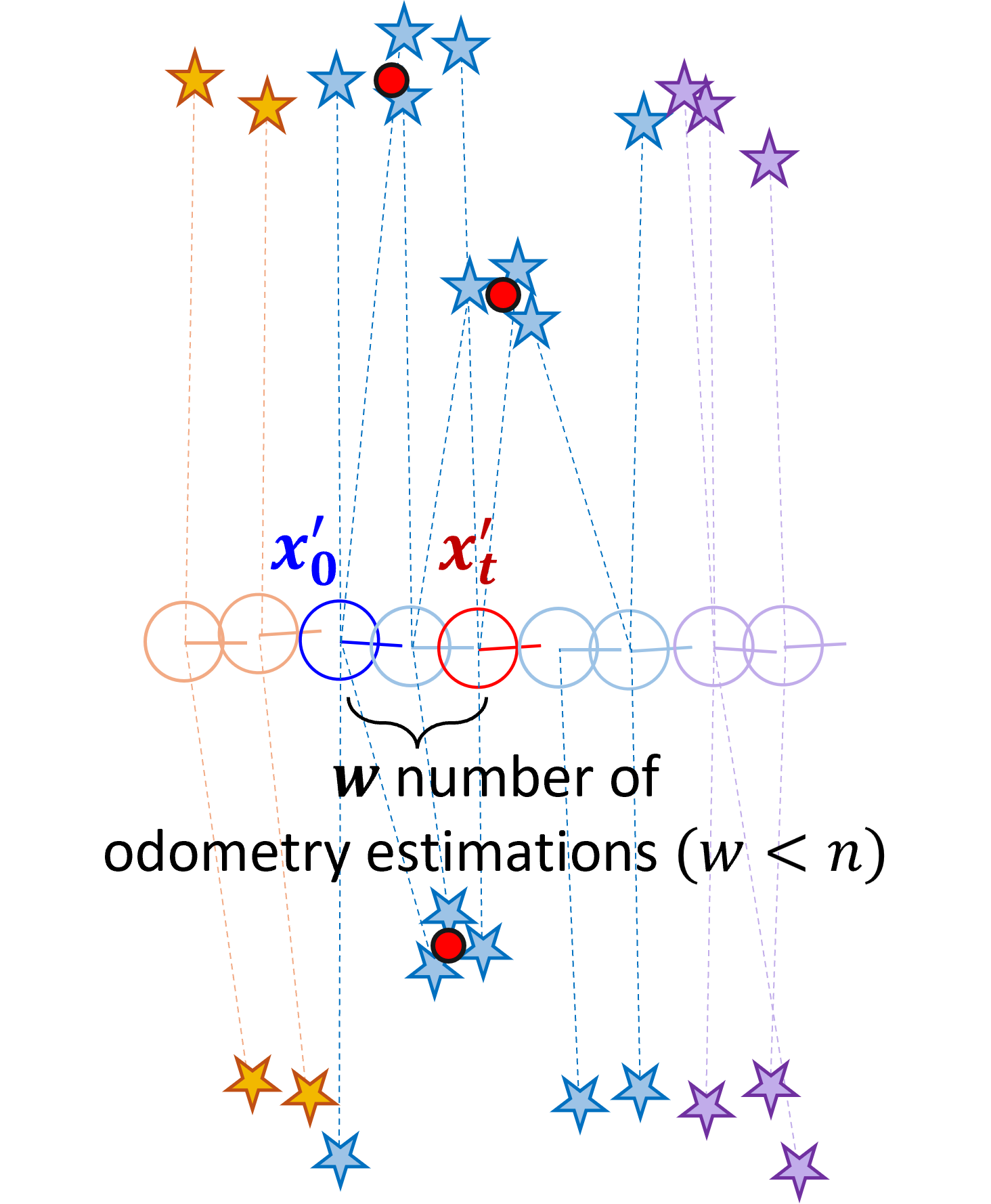}%
		\label{step2}}
	\hfil
	\subfloat[]{\includegraphics[height=1.75in]{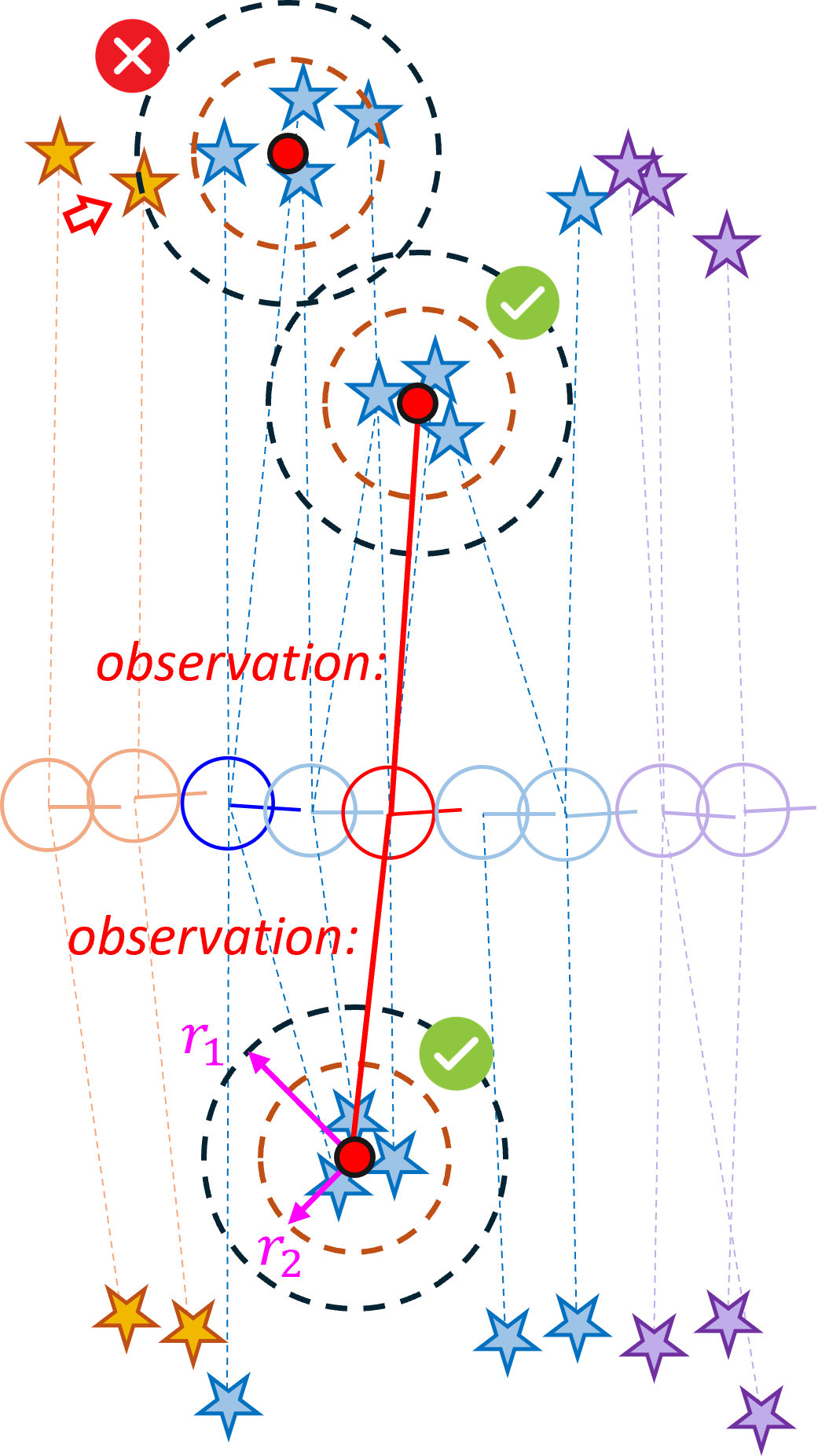}%
		\label{step3}}
	\hfil
	\subfloat[]{\includegraphics[height=1.75in]{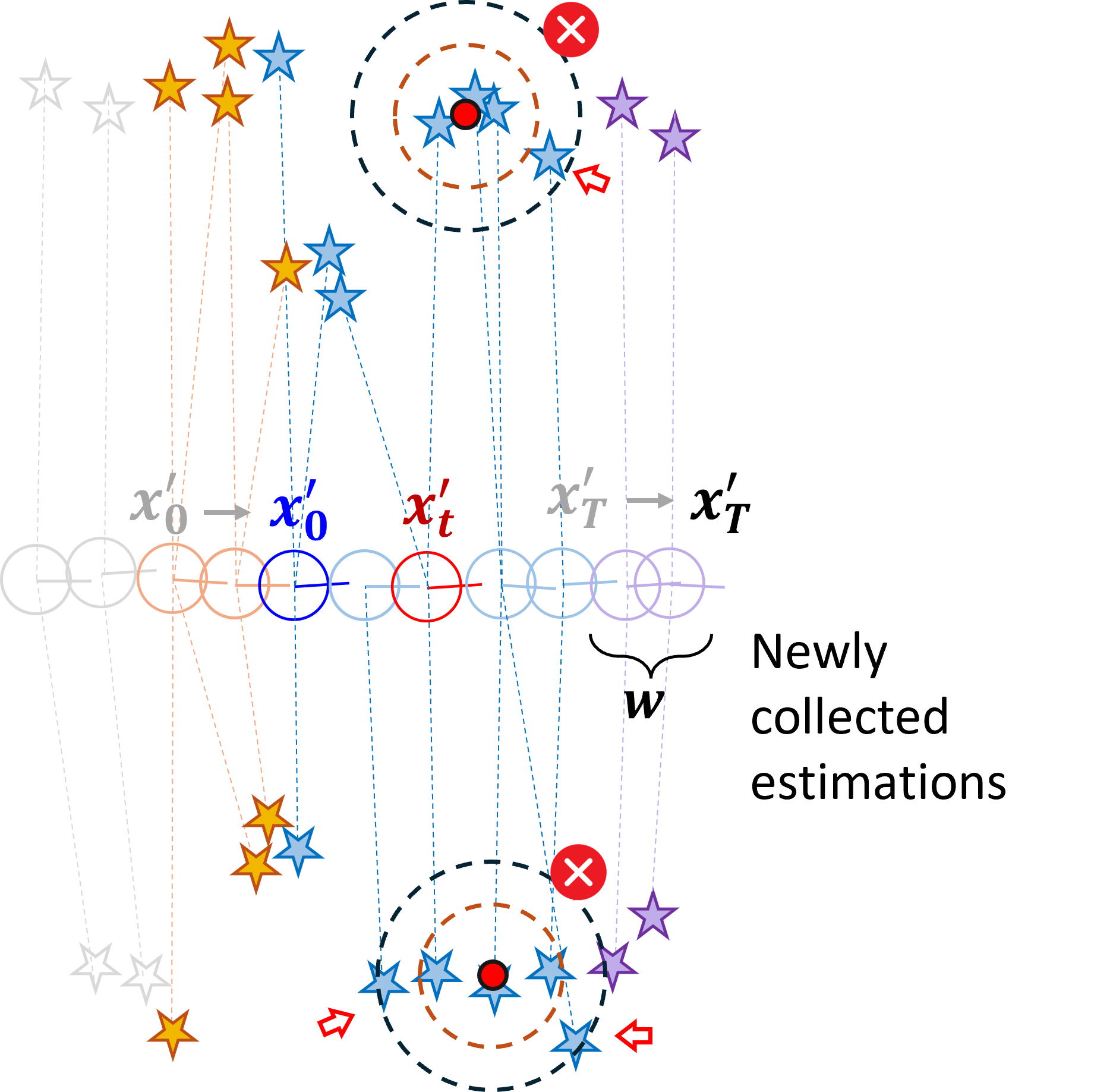}%
		\label{step4}}
	\caption{\textbf{Point feature extraction steps using UWB radar.} a) Accumulation of a set of provisional observations with respect to $m_1 + n + m_2$ number of odometry pose estimations and create a point cloud. The region of interest to extract point features lies in the middle with $n$ estimations. The surrounding history and some incoming $m_1 + m_2$ number of observations are also accumulated which will be useful to isolate the points. b) The point cloud is clustered using DBSCAN. Mean of each cluster is sent to the next step. c) Count the number of points inside two circles with radii $r_1$ and $r_2 (< r_1)$. If the number of points inside both circles are equal, that particular mean is considered as an observation from a prominent point feature. d) The robot proceeds to the next update step by accumulating new observations while discarding
		past provisional observations in a moving window fashion.}
	\label{fig_sim}
\end{figure*}

In the context of UWB AOA anchor-tag systems, Wang et al. \cite{wang} developed an UWB antenna array with Decawave UWB ICs to obtain AOA information. 
They have localized several agents using a single static anchor under line-of-sight (LOS) conditions (i.e. omitting multi-path propagation). 
Similarly, Zhong et al. \cite{aoa_1}  proposed a UWB AOA-based indoor position tracking system. A UWB AOA anchor had been mounted on a rotating platform that tracks a moving tag, and a least squares-based filtering algorithm was proposed to eliminate 
anomalous data caused by occlusions.

\textcolor{black}{On the other hand, UWB range-only TOF readings are frequently being used in SLAM systems with multiple UWB beacons \cite{ROSLAM, liu2024rangeslamultrawidebandbasedsmokeresistantrealtime}.}
These range-only measurements are often fused with another exteroceptive sensor (e.g. LiDAR) to improve SLAM performance \cite{uwblidar, radio}. 
In this aspect, Zhang et al. \cite{RN379} used state-of-the-art \textit{LinkTrack UWB AOA} anchor-tags by \textit{Nooploop} to fuse visual inertial measurements with UWB AOA readings. 
Although they propose statistical methods to filter outliers in UWB AOA readings, their SLAM experiments use simulated AOA measurements, thus omitting multi-path propagation in complex real-world environments. 
However, they have demonstrated the potential of UWB AOA anchor-tag systems in the SLAM domain.

In this paper, we propose All-UWB radar SLAM, a novel approach fusing UWB AOA information with existing landmark-based UWB radar SLAM for feature-deficient unstructured indoor environments. We deploy artificial landmarks (i.e. UWB AOA nodes) in feature-deficient areas to mitigate error accumulation during SLAM. \textcolor{black}{Dynamic deployment is preferred over static nodes due to less infrastructure dependency, robustness and adaptive spatial coverage \cite{sensorfly}. Moreover, compared to range-only systems \cite{ROSLAM} and fingerprinting approaches with beacons \cite{drunkwalk}, an AOA anchor-tag pair provides both range and bearing measurements, serving as independent artificial landmarks.}

This paper thoroughly discusses solutions to the challenges in post-processing UWB AOA observations instead of diving deep into the calculations behind UWB AOA estimations. 
Hence, the authors believe that this work will facilitate researchers and engineers to improve their SLAM systems by exploiting UWB AOA observations. Lastly, we share our datasets and the ROS2 implementation of our UWB SLAM system with the research community.\footnote{\href{https://github.com/CPrem95/ALL\_UWB\_SLAM.git}{https://github.com/CPrem95/ALL\_UWB\_SLAM.git}}

\section{Proposed Approach} \label{sec::prop_appr}

\subsection{System Overview} 

Fig. \ref{oview} illustrates the overview of the proposed system. The frontend mainly comprises two types of sensors: UWB radar and UWB AOA anchor-tags. The plan view of the sensor configuration is depicted in Fig. \ref{sensor_config} for a non-holonomic robot. The proposed setup can be further extended to a holonomic robot by using UWB radar sensor arrays to cover the entire 360$^\circ$ view.
The UWB radar arrays find point features in the environment while the AOA anchors get observations from the isolated tags. A moving window-based approach is followed to filter anomalous observations.
In addition, wheel odometry is fed to the SLAM backend as robot motion updates.

Landmark-based vanilla UWB radar SLAM systems generally assume that the environment has sufficient features, and is at least partially structured with low to zero clutter. However, this assumption is not applicable to most real-world scenarios. Hence, this paper proposes two main solutions to address this issue: 
1) Extract only the prominent features in the environment by isolating only the prominent point features in the environment; 
2) Artificially deploy landmarks in feature-deficient areas to mitigate odometry drift and facilitate loop closing. These artificial landmarks
consists of UWB AOA tags, which will then be detected by AOA anchors of the robot. Hence, as shown in Fig. \ref{oview}, the frontend mainly employs UWB technology (UWB radar and AOA anchor-tags) to perceive the surroundings as  exteroceptive sensors. Accordingly, we designate the proposed system as All-UWB SLAM.

\subsection{UWB Radar} \label{II-C}
The UWB radar arrays comprise Novelda X4M300 modules with directional antennas. Their overlapping FOVs enable trilateration \cite{cp1}.

\subsubsection{Signal Preprocessing} \label{sect_sp}
UWB radar observation is a time-series, where the x-axis corresponds to the time delay between signal transmission and the reception of radar reflections (see Fig. \ref{raw_signal}). This delay provides information about the distance to objects in the radar's field of view. High amplitudes indicate a high likelihood of an object with a large radar cross-section (RCS). We use a UWB radar array to obtain scan points of the surroundings (see Fig. \ref{sensor_config}).

The preprocessing steps are as follows:
{
\renewcommand{\labelenumi}{\alph{enumi})}
\begin{enumerate} 
	\item Take the absolute values of the amplitudes.
	\item Smoothen the signal (apply a Savitzky-Golay filter to preserve the peaks).
	\item Find the peaks by defining suitable thresholds.
	\item Get the distance to the target by converting the corresponding peak timestamp from time domain to the distance domain.
	\item Trilaterate the objects that are within the overlapping region of the radar's directional antennas.
\end{enumerate}
}
\subsubsection{Feature Extraction} \label{3C2}
The trilaterated observations are accumulated with respect to the odometry pose estimations, and point features are extracted in a moving window fashion (see Fig. \ref{fig_sim}). Whenever the odometry exceeds a predefined displacement threshold \textit{min\_disp}, the robot gets scan points of the surroundings by preprocessing the current UWB radar observations. 
After $n$ number of such instances, the robot has accumulated a set of provisional observations w.r.t. the odometry estimations (see Fig. \ref{fig_sim}a). 
We use this accumulated point cloud to extract point features in the environment. The initial pose $x_0^{'}$ and the pose after $w (<n)$ instances $x_t^{'}$ are used at the backend to predict the next state using the odometry motion model. 

UWB radar detects point features in the environment as densely clustered scan points.
Hence, we apply DBSCAN to the accumulated point cloud to identify clusters. The mean values of those cluster centers are sent to the next step as potential point feature candidates (see Fig. \ref{fig_sim}b).
Meanwhile, the robot keeps scans from $m_1$ previous and $m_2$ incoming instances to isolate only the prominent point features.

\begin{figure}[!t]
	\centering
	{\includegraphics[height=1.77in]{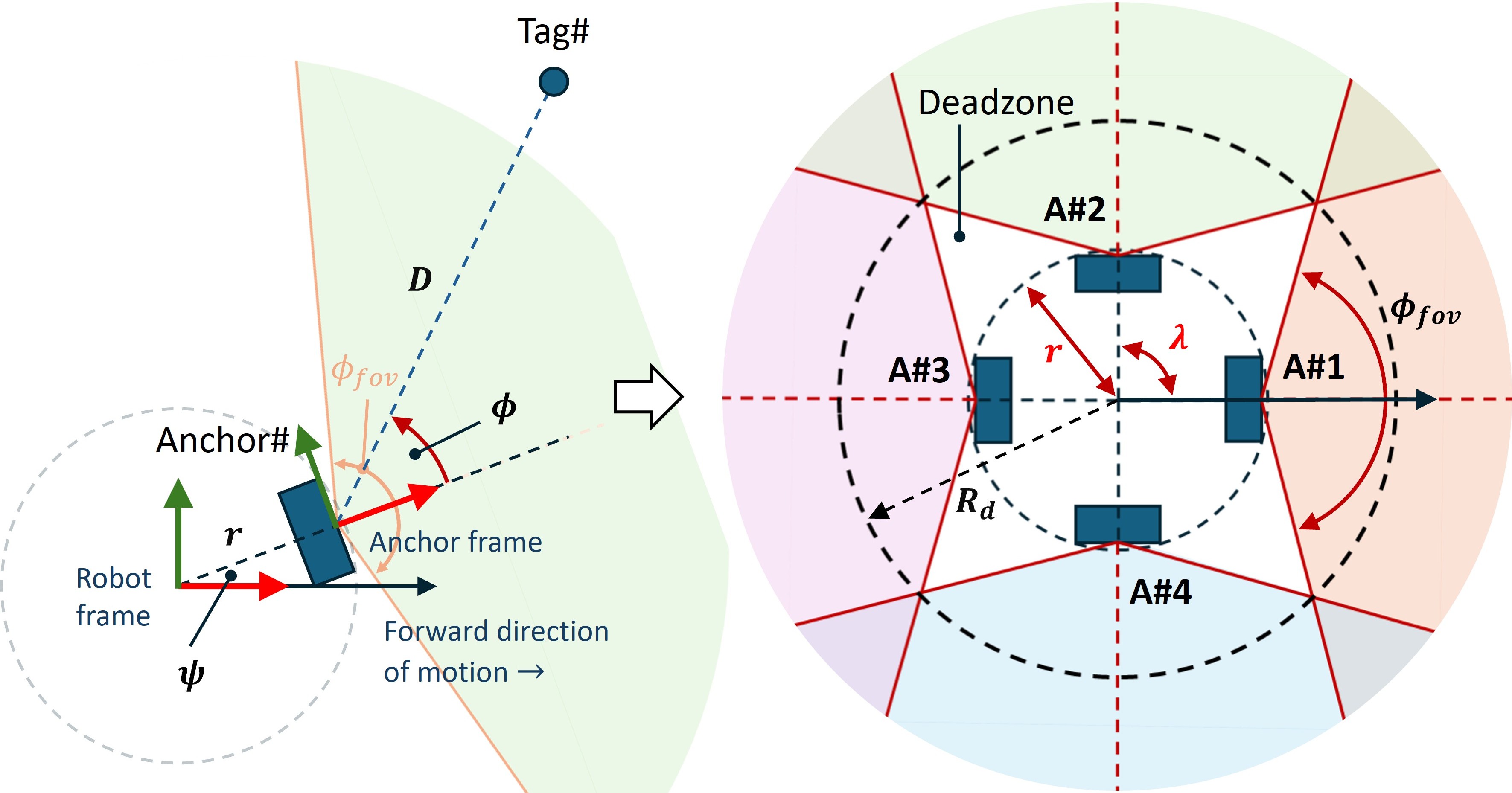}}
	\caption{\textbf{Proposed anchor configuration using LinkTrack AOA anchors.} The azimuthal FOV of the anchors ($\phi_{fov}$) was considered when designing the sensor array. Variables required to transform the Tag observations to the robot frame (left), and the final anchor configuration (right). Notice that there are several deadzones within a radius $R_d$ which should be minimized.}
	\label{ancs}
\end{figure}

\subsubsection{Isolating Prominent Point Features} \label{sect_ippf}
The clustered observations may belong to a cluttered area in the environment, thus may not belong to a point feature. Initialising such features as point landmarks may lead to degradation of the SLAM results. Hence, we count the number of scan points within two circles with radii $r_1$ and $r_2 (<r_1)$ centered at the cluster mean. If both counts are equal, that is registered as a prominent point feature. This is when the $m_1 + m_2$ number of early and incoming scan instances are useful (see Fig. \ref{fig_sim}c). Finally, the range and bearing observations of the identified point features (w.r.t. the $x_t^{'}$) are sent to the SLAM backend. After that, the robot collects more scan points in a moving window and updates the SLAM results (see Fig. \ref{fig_sim}d).

\subsection{UWB AOA Anchor-tag Modules} \label{IID}

We have used LinkTrack AOA modules by Nooploop to illustrate the concept of integrating UWB AOA anchor-tags. The anchor (i.e. base station) determines the range and bearing (i.e. AOA readings) to a tag relative to itself, not vice versa \textcolor{black}{(see Fig. \ref{ancs})}. Hence, if we arbitrary put anchors in unknown locations in the environment, the SLAM system has to determine the orientation of the anchors (i.e. artificial landmark) in addition to the position. This increases the dimensionality of the SLAM problem, leading to higher computational costs. Moreover, due to the $\pm5^\circ$ specified accuracy of the sensor~\cite{aoa} (noting that the empirical accuracy is lower than the given specifications), the artificial landmark's observations may not significantly improve the robot's pose correction until its orientation uncertainty is substantially reduced.

Considering above facts, the LinkTrack tags are deployed in the environment instead of anchors. The anchors are mounted on the robot enabling to obtain range and bearing observations of the tags relativ to the robot's frame. The tags are deployed in feature-deficient areas while exploring the environment.
Generally, UWB ranging-based anchor-tag sytems require at least three noncollinear anchors to localize the tag in 2D. Mathematically, only two measurements are enough to solve for the 2D coordinates, despite the mirror ambiguity with two range readings. In the proposed setup, we obtain both range and bearing information of the deployed tags relative to the robot. Therefore, we use the term: `isolated tags' to imply that the tags can function independently without being discovered by the peers. This is critical when exploring a complex environment, since NLOS conditions significantly reduce the accuracy of UWB measurements.

\begin{figure}[!t]
	\centering
	{\includegraphics[width=2.7in]{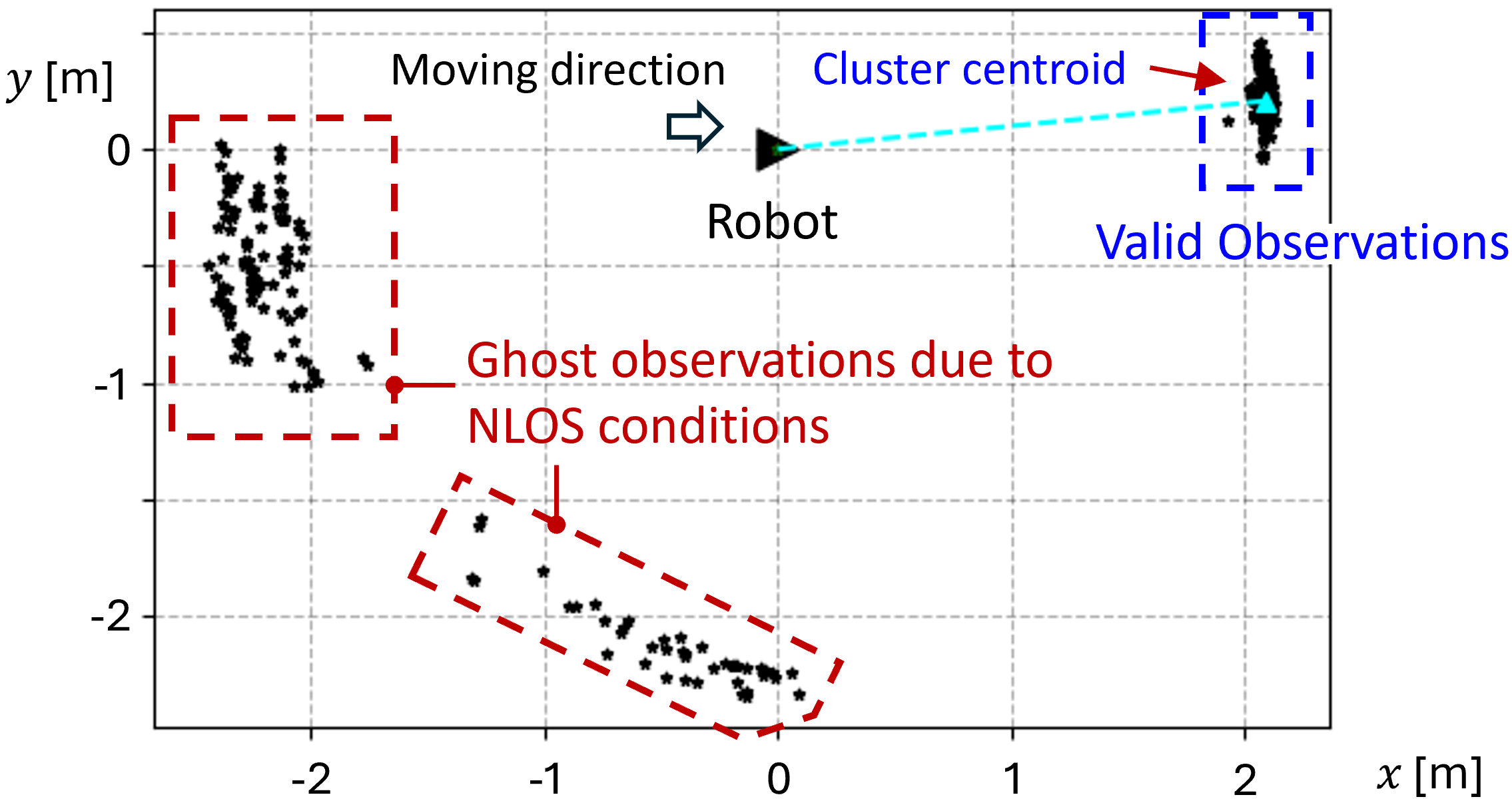}}
	\caption{\textbf{Filtering of the ghost observations from UWB AOA modules.} The robot tends to observe ghost observations from the deployed tags in the environment. Here, we have exploited the fact that the robot is moving while performing SLAM in order to omit ghost observations (dispersed) from the valid observations (clustered). The anchor ring occasionally detects multiple ghost observations since more than one anchor is susceptible to multi-path propagation and front-back ambiguities. }
	\label{ghost}
\end{figure}

\subsubsection{Anchor Ring Setup}\label{IIIC1}
There exists $N_A$ number of equi-spaced anchors ($\lambda$) creating a circle of radius $r$ centred around the robot's frame origin. According to the Fig. \ref{ancs}, the AOA anchor's range ($D$) and bearing ($\phi$) observation can be transformed to the robot's frame in Cartesian coordinates $[X_R, Y_R]$ using the following equations:
\begin{equation}
T(\theta_Z, D_X) = 
\begin{bmatrix}
	\cos(\theta_Z) & -\sin(\theta_Z) & D\cos(\theta_Z) \\
	\sin(\theta_Z) & \cos(\theta_Z) & D\sin(\theta_Z) \\
	0 & 0 & 1
\end{bmatrix}
\end{equation}

\begin{equation}
\begin{bmatrix}
	X_R \\
	Y_R \\
	1
\end{bmatrix}
=
T(\psi, r).T(\phi, D).
\begin{bmatrix}
	0 \\
	0 \\
	1
\end{bmatrix}
\end{equation}

The radius of the deadzone $R_d$ \textcolor{black}{(refer Fig. \ref{ancs})} can be found using the following equation:

\begin{equation}
R_d = \frac{r.\tan(\phi_{fov}/2)}{\tan(\phi_{fov}/2).\cos{(\lambda/2)} - \sin{(\lambda/2)}}
\end{equation}

where $\phi_{fov}$  is the azimuthal FOV of LinkTrack AOA anchors. We have used $N_A =$ 4 anchors, thus $\lambda = 360^\circ/N_A$ $ = 90^\circ$, and $\psi$ values of each anchor [A\#1, A\#2, A\#3, A\#4] is given by $\psi = [0^\circ, 90^\circ, 180^\circ, -90^\circ]$. Our $r =$  10 cm, and $\phi_{fov} = $ 150$^\circ$ leads to $R_d =$ 87.8 cm ($<$ 1 m). Notice that $R_d$ can be minimized by using a small $r$, and by adding more anchors (i.e. reducing $\lambda$). 

\subsubsection{Identifying a Feature-deficient Region}\label{sect_fda} 
The features in the environment are identified at the frontend by the UWB radars (refer Section \ref{II-C}). Whenever the robot observes a new or existing feature, it memorises the corresponding pose estimation. After the robot has travelled a certain distance threshold (\textit{dep\_dist}) without observing any feature, the robot identifies the surroundings as a potential feature-deficient area, and a UWB tag can be deployed.

\begin{figure}[!t]
\centering
{\includegraphics[clip,trim=0 0in 0 0,  width=2.8in]{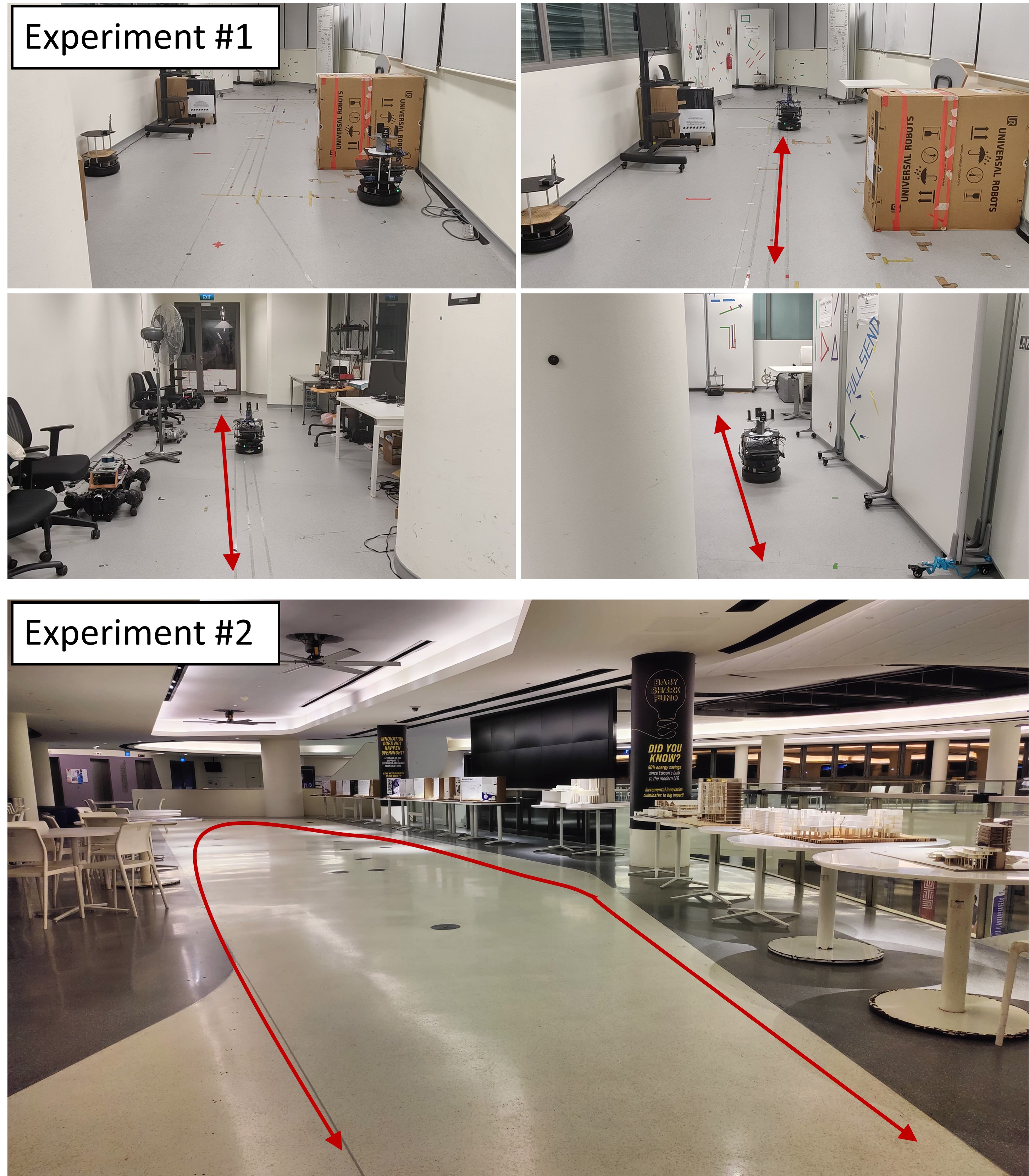}}
\caption{\textcolor{black}{\textbf{Experimental environments.} The experiments were conducted in two different unstructured indoor environments comprising only a few prominent features. The mobile robot (refer Fig. \ref{sensor_config} for the sensor configuration) was teleoperated as indicated by the arrows.}}
\label{env}
\end{figure}

\subsubsection{Initializing the Tags}

\textcolor{black}{Upon deployment of a tag in the environment, the robot halts and initially acquires multiple observations from the tag.}
Then, the circular mean $[\mu_r, \mu_\theta]$ and standard deviation $[\sigma_r, \sigma_\theta]$ of the observations are calculated.
\begin{equation}
\resizebox{0.9\hsize}{!}{$
	\bar{x} = \frac{1}{n} \sum_{i=1}^n \cos\left(\theta_i\right), \text{ }
	\bar{y} = \frac{1}{n} \sum_{i=1}^n \sin\left(\theta_i\right), \text{ }
	R = \sqrt{\bar{x}^2 + \bar{y}^2}
	$}
	\end{equation}
	\begin{equation}
\mu_\theta = \operatorname{atan2}\left(\bar{y}, \bar{x}\right), \quad
\sigma_\theta = \sqrt{-2 \ln(R)}.
\end{equation}
Only the points that exist within $\mu \pm \sigma$ are considered. Finally, the tag is initialized using the mean of the filtered observations with respect to the current pose estimation.

\subsubsection{Acquiring Observations} \label{sect_ao}

\textcolor{black}{Similar to the accumulated point cloud using UWB radar, we collect AOA observations from the tags relative to the robot. 
However, under NLOS conditions resulting from multi-path propagation and front-back ambiguities \cite{angu}, the anchors tend to observe ghost tags in unusual locations that significantly deviate from the actual location (see Fig. \ref{ghost}).
Hence, we collect AOA observations only if the robot is moving (i.e. robot's velocity $>$ \textit{min\_vel} $>$ 0). 
As a result, the ghost points get scattered when the robot traverses areas that causes such anomalous observations. 
Finally, the valid clustered observations are extracted 
using DBSCAN and the cluster centroid is registered as the observation.}
\textcolor{black}{
In the rare event of ghost points forming one or more clusters, they are treated as outliers. These outliers are completely discarded by checking if the number of clusters $> 1$, or if the Mahalanobis distance between the observation and the current tag estimation (i.e. landmark) exceeds a predefined threshold ($\alpha_t$).
}

\begin{algorithm}[t]
\definecolor{red(pigment)}{rgb}{0.93, 0.11, 0.14}
\definecolor{black(pigment)}{rgb}{0, 0, 0}
\color{black(pigment)}
\caption{All-UWB SLAM Backend}\label{startAlgo}
\hspace*{\algorithmicindent} \textbf{Input:} Robot's motion uncertainty $R$, observation uncertainty of point features $Q_r$ and tags $Q_t$, displacement threshold to accumulate an observation \textit{min\_disp}, moving window size: $m_1 + n + m_2$, $w$ (ref Fig. \ref{fig_sim}).
EKF SLAM: initial mean $\mu$, initial covariance $\Sigma$, Mahalanobis dist thresholds ($\alpha_r$, $\alpha_t$), odometry data, observations: UWB radar and tags.
\newline
\hspace*{\algorithmicindent} \textbf{Output:} robot pose and covariance.

\begin{algorithmic}[1]
	\color{black(pigment)}
	\State $s \gets 1$\Comment{id of the estimated state}
	\State $i \gets 1$\Comment{id of the current state}
	\State $\text{Rad\_Obs[]} \gets \text{empty list of UWB radar observations}$
	\State $\text{Tag\_Obs[]} \gets \text{empty list of tag observations}$
	\State $\text{Odom[]} \gets \text{empty list of odometry}$
	\State  $\textit{previous pose} \gets \text{read odometry}$\Comment{ $\text{Odom[0]} \gets \text{origin}$}
	
	\While{\textit{teleop}} \Comment{do while teleoperating the robot}
	\State $\textit{new pose} \gets \text{read odometry}$
	\State $\textit{disp} \gets \textit{new pose} - \textit{previous pose}$
	\If {$ \text{\textit{robot\_velocity}} > \textit{min\_vel}$}
	\State $\text{Tag\_Obs[]} \gets \text{observations from UWB AOA tags}$
	\EndIf
	\If {$ \text{\textit{disp}} > \textit{min\_disp}$}
	\State $\text{Rad\_Obs[]} \gets \text{observations from UWB radar}$ 
	\State Odom[$i$] $ \gets \textit{new pose}$
	\State $i \gets i + 1$
	
	\If{$i > n + m_2$} \Comment{executes after collecting initial set of observations from $n + m_2$ number of poses}
	\
	\State \textit{filt\_pts} $\gets\text{\textit{filter}(Rad\_Obs)}$ \Comment{Section \ref{3C2}}
	\State \textit{filt\_tags} $\gets\text{\textit{filter}(Tag\_Obs)}$ \Comment{Section \ref{sect_ao}}
	\State \textit{u} $\gets$ odometry motion model
	\Statex	\Comment{calculated using Odom[$s$] and Odom[$s + w$]}
	\State $\bar{\mu}, \bar{\Sigma} \gets$ EKF prediction$(\mu, \Sigma, \textit{u}, \textit{R})$
	\State $\mu, \Sigma \gets$ EKF update$(\bar{\mu}, \bar{\Sigma}, Q_r, \alpha_r, \text{\textit{filt\_pts}})$
	\State $\mu, \Sigma \gets$ EKF update$({\mu}, {\Sigma}, Q_t, \alpha_t, \text{\textit{filt\_tags}})$
	\State $s \gets s + 1$
	\EndIf
	\State Keep observations from $m_1 + n + m_2$ instances and ignore the rest \Comment{moving window}
	\State $\textit{previous\_pose} \gets \textit{new\_pose}$
	\State\Return{robot pose $\gets \mu, \Sigma$}
	\EndIf
	\EndWhile
\end{algorithmic}
\label{algo1}
\begin{tablenotes}
	\footnotesize
	\vspace{-0.5em}
	\item \textbf{\textit{Note:}} This pseudocode has intentionally omitted tag deploying steps. Ignore the tag update step if it has not yet been initialized/deployed.
\end{tablenotes}
\end{algorithm}

\subsubsection{Key Insights from the Developer's Documentation}
The anchors and tags have a field of view of 150$^\circ$. This value was taken into consideration when designing the configuration of the anchor array. The anchor-tag units should be at least 0.5~m higher from the ground and the anchors should be at least 10 cm away from walls to avoid noisy observations. 
The developers suggest to find the difference between the first-path received signal strength indicator (RSSI: $fp\_rssi$) and the total RSSI ($rx\_rssi$) to determine LOS conditions.
\begin{equation}
\Delta_{rssi} = rx\_rssi - fp\_rssi 
\end{equation}
If $\Delta_{rssi} < 6$ dB, the tag is likely to be in LOS state and vice versa. However, we can reduce the RSSI threshold (6 dB) considering factors  such as environmental conditions (dynamic/static), and the expected reliability of the observations. 

\subsection{SLAM Backend} \label{sect_sb}

We have used EKF SLAM at the backend to illustrate the performance of the proposed approach (refer Algorithm \ref{algo1}). Since the anchors know the IDs of the observed tags, the AOA observations are fed to the update step of EKF SLAM with \textit{known} correspondences. However, the identity of the features extracted from the UWB radar are still unknown. Hence, we have used a nearest neighbor approach using the squared Mahalanobis distance (M-distance) to solve the data association problem for \textit{unknown} correspondences. Whenever the UWB radar receives a new observation, it is associated to an existing landmark that has the minimum M-distance from the new observation. If the minimum M-distance exceeds a predefined threshold ($\alpha_r$), the observation is initialized as a new landmark.

\section{Experiments and Results} \label{sec::exp_and_res}

\begin{figure}[!t]
	\centering
	{\includegraphics[clip,trim=0in 0in 0in 0, width=3.4in]{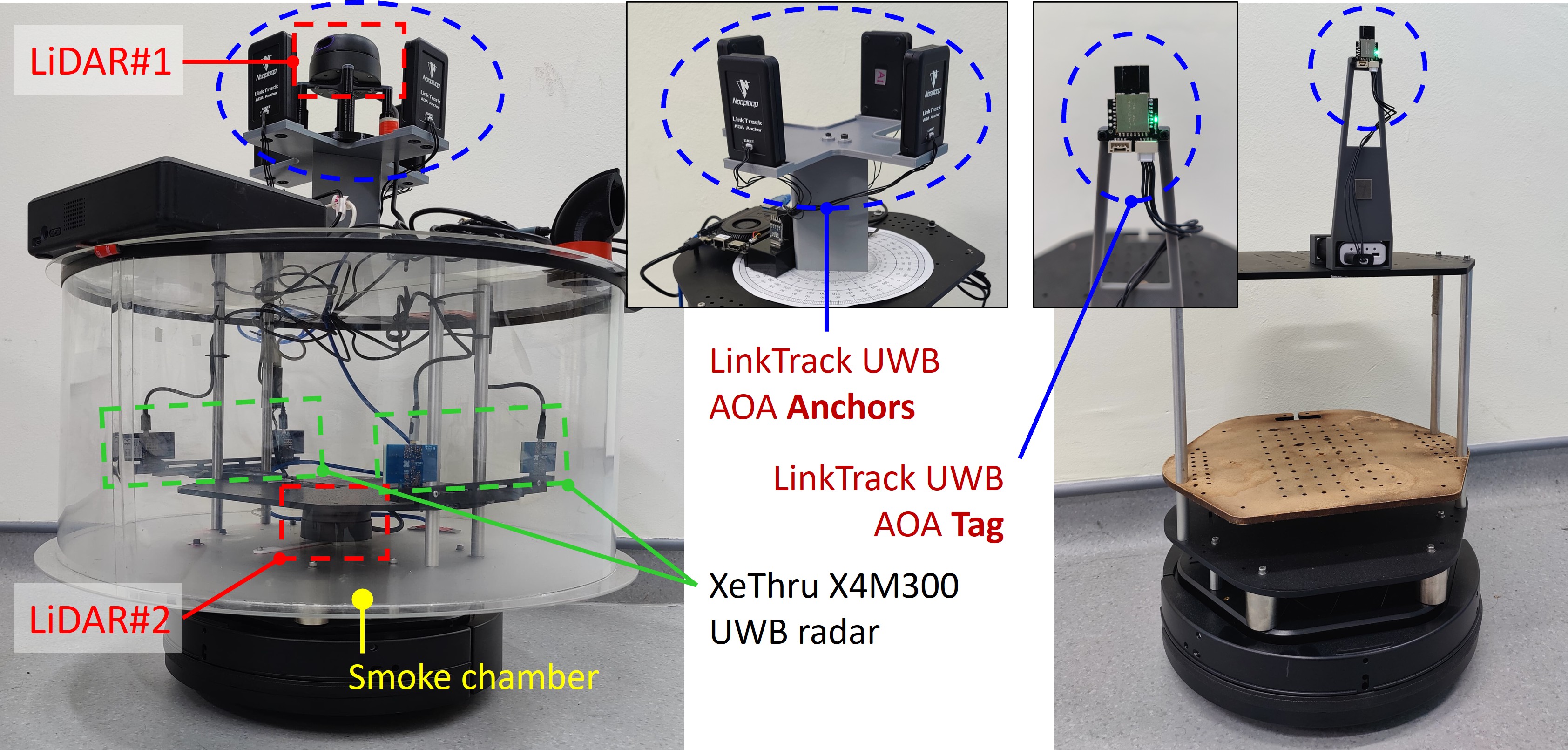}}
	\caption{\textbf{The mobile robot with the sensing setup.} A non-holonomic mobile robot is used: TurtleBot2 (\textit{left}). The AOA anchors (Nooploop's LinkTrack AOA) are mounted on top and the UWB radar modules (Novelda's X4M300) are mounted on both sides. \textcolor{black}{The LiDAR\#1 is used to obtain the ground truth, whereas the LiDAR\#2 tries to perceive through smoke.} 
		Another TurtleBot(s) is used to emulate tag deployment (\textit{right}). The tags do not require processing units; a suitable power supply is sufficient.}
	\label{tbot}
\end{figure}

\begin{figure}[t]
	\centering
	{\includegraphics[height=1.8in]{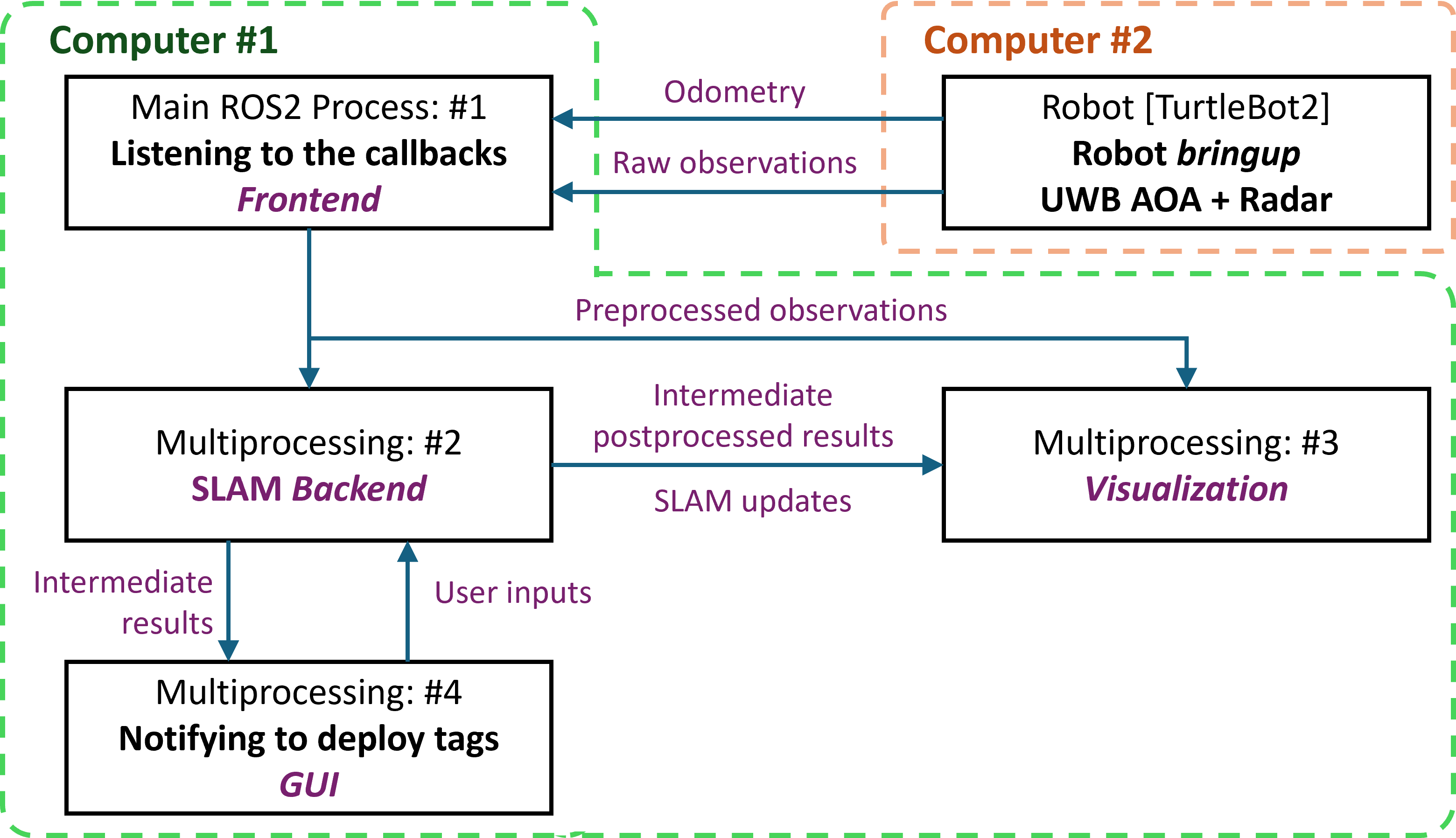}}
	\caption{\textbf{Visual representation of the software architecture.} The SLAM system was implemented with ROS2 middleware. 
		All components run concurrently in separate processes (Python \textit{multiprocessing}), enabling real-time capture of sensor observations and processing them independently.
	}
	\label{program}
\end{figure}

To evaluate the proposed All-UWB SLAM system, experiments were conducted in \textcolor{black}{two indoor environments with a size of 12 × 18 m$^\text{2}$ and 12 × 26 m$^\text{2}$ (see Fig. \ref{env}).} These unstructured environments comprised of a few objects with point features (e.g. table legs, fan) and clutter (e.g. robots, partition walls).
Fig. \ref{tbot} illustrates the configuration of the sensors (i.e. UWB radar, UWB AOA anchors and \textcolor{black}{LiDARs) mounted on a TurtleBot. The TurtleBot is contained inside a smoke chamber to emulate a vision-denied scenario.}
There are two computers onboard: 1) a LattePanda Delta to run \textit{TurtleBot2\_bringup} and to get UWB AOA observations from the anchors; 2) a laptop with an i7 processor to get UWB radar observations and LiDAR scans.
Another workstation laptop (11th gen Intel Core i9) is connected to the robot via WiFi. It executes the SLAM backend within ROS2 framework from the operator's side (see Fig. \ref{program}).

\begin{figure}[!t]
	\centering
	{\includegraphics[clip,trim=0 0in 0 0, width=3.45in]{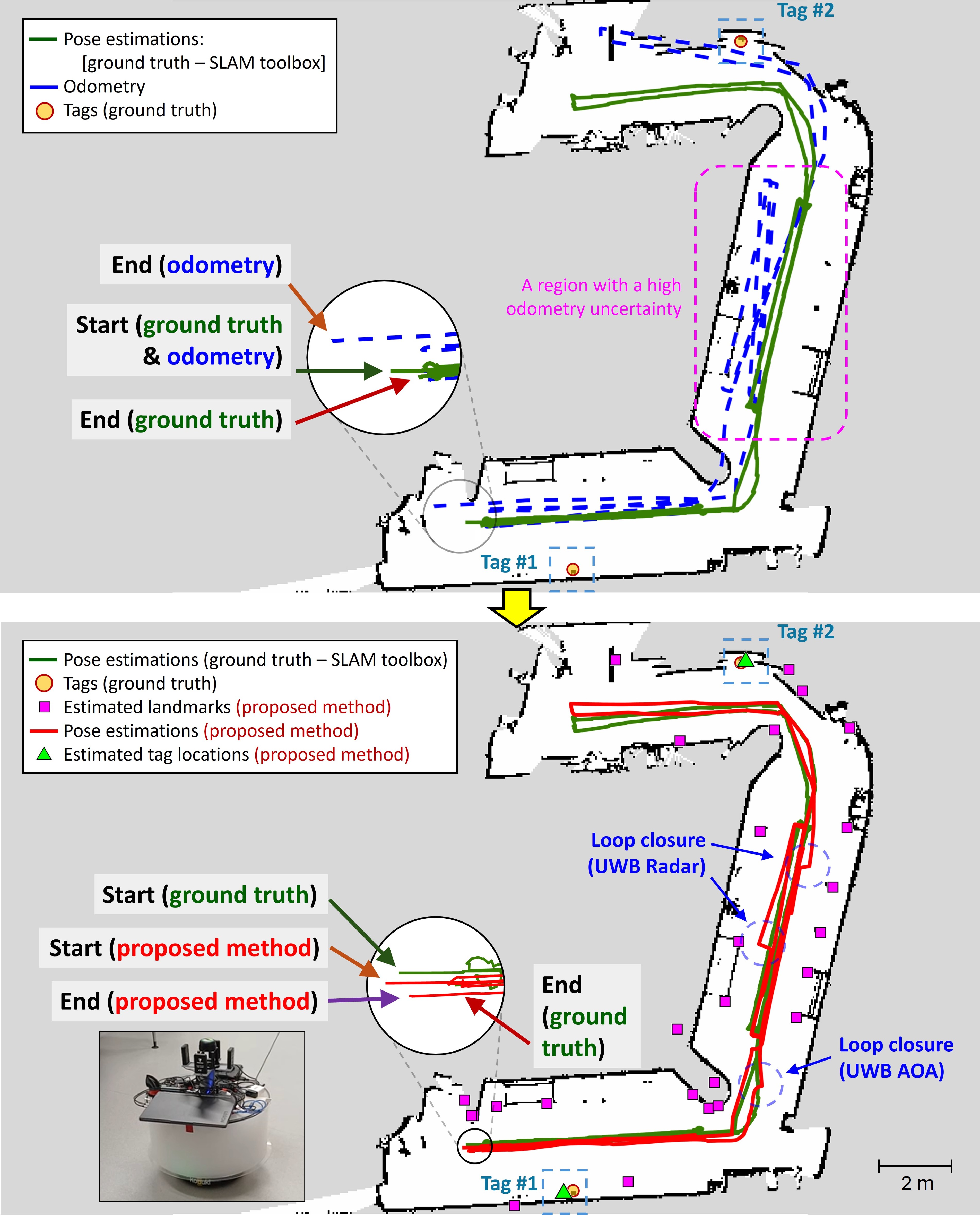}}
	\caption{\textcolor{black}{\textbf{Experiment \#1: All-UWB SLAM in a vision-denied environment.} The smoke chamber was filled with smoke and the robot was teleoperated in a U-shaped path in an indoor environment.
			Notice that the wheel odometry has significantly deviated from the ground truth (top). The landmark map of the proposed All-UWB SLAM system is overlaid on the occupancy grid map from the SLAM toolbox as ground truth (bottom).}
	}
	\label{ex_lab}
\end{figure}

\begin{figure}[!t]
	\centering
	{\includegraphics[clip,trim=0 0in 0 0, width=3.45in]{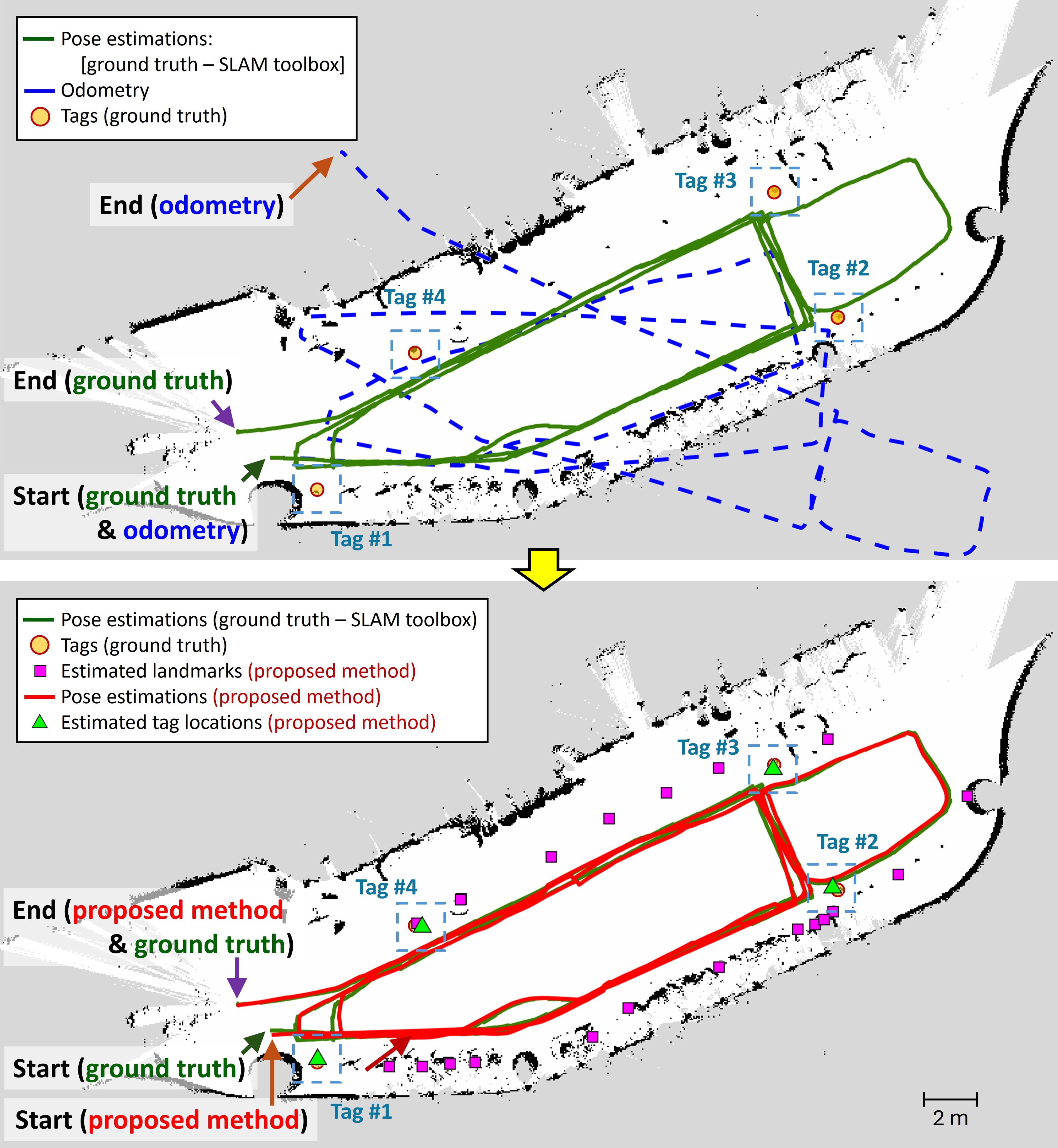}}
	\caption{\textcolor{black}{\textbf{Experiment \#2: All-UWB SLAM in a larger environment.} Four tags were deployed while exploring the environment under high wheel odometry uncertainty (top). The generated landmark map from the proposed system was overlaid on the ground truth from the SLAM toolbox (bottom). }
	}
	\label{ex_hall}
\end{figure}

\begin{table}[t]
	\caption{Important parameters used in the proposed All-UWB SLAM}\centering
	\begin{center}
		\setlength{\extrarowheight}{1pt}%
		\begin{tabular}{|p{0.07\textwidth}|p{0.085\textwidth}|p{0.255\textwidth}|}
			\hline
			\textbf{Parameter}&\textbf{Range}&\textbf{Remarks}\\
			\hline
			$n, m_1, m_2$ \newline $w$ & $50, 150, 50$ \newline $2$ & Parameters of the moving window, and \newline the step size for SLAM update. \newline Refer Section \ref{3C2} and Fig. \ref{fig_sim}. \\
			\hline
			$\textit{min\_disp}$& $5$ mm \textit{or}\newline 
			\resizebox{1.1\hsize}{!}
			{$2.5 \times 10^{-3}$ \text{ rad\newline }} & Updates the provisional observations list once the robot moves exceeding these min displacement thresholds.\\
			\hline
			$N_A$& $4$ & Number of anchors on the robot. \\
			\hline
			$r$& $10$ cm & Radius of the anchor ring.\\
			\hline
			$\textit{det\_range}$& $1.5 - 10$ m & UWB AOA detection range.\\
			\hline
			$\textit{min\_vel}$&$3$ cm/s \textit{or} \newline $0.01$ rad/s & Minimum velocity to obtain UWB \newline AOA readings (refer Section \ref{sect_ao}).\\
			\hline
			$\textit{dep\_dist}$&$1$ m & Min travel distance to identify a feature-deficient region (Section \ref{sect_fda}).\\
			\hline
			$[\textit{eps, n}]_{t}$& $[5\text{ cm}, 10]$& DBSCAN-based filtering for AOA \newline readings: max distance between two observations (\textit{Euclidean}) and the min samples per cluster (Section \ref{sect_ao}).\\
			\hline
			$[r_1, r_2]$& $[30, 15] \text{ cm}$ & Radii of the circles isolating the point feature. Refer Sec. \ref{sect_ippf} and Fig. \ref{fig_sim}.\\
			\hline
			$[\textit{eps, n}]_{r}$& $[20 \text{ cm}, 10]$& DBSCAN-based filtering for AOA \newline readings: max distance between two observations (\textit{Euclidean}) and the min samples per cluster (Section \ref{sect_ao}).\\
			\hline

			$R$& \text{ }
			\newline 
			\resizebox{0.09\textwidth}{!}
			{$
				\begin{bmatrix}
					\sigma_x^2 & 0 & 0 \\
					0 & \sigma_y^2 & 0\\
					0 & 0 & \sigma_\theta^2\\
				\end{bmatrix}
				$}
			
			& Motion uncertainty of the robot between two poses. Depends on \textit{min\_disp}. Larger the \textit{min\_disp} and $w$ is, larger the uncertainty will be, and vice versa. \newline
			\textcolor{black}{$\sigma_x = \sigma_y = 1$ mm \newline
				$\sigma_\theta = 5 \times 10^{-3}$ rad}  \\
			\hline
			$Q_{t}$& \text{ }
			\newline 
			\resizebox{0.06\textwidth}{!}
			{$
				\begin{bmatrix}
					\sigma_r^2 & 0\\
					0 & \sigma_\phi^2\\
				\end{bmatrix}
				$}
			& Observation uncertainty. Large uncertainty due to inherent measurement\newline noise. \textcolor{black}{{$\sigma_r \text{ = } 30\text{ cm, } \sigma_\phi^2 \text{ = } 1 \text{ rad$^2$\newline }$}}\\
			\hline
			$Q_{r}$& \text{ }
			\newline 
			\resizebox{0.06\textwidth}{!}
			{$
				\begin{bmatrix}
					\sigma_r^2 & 0\\
					0 & \sigma_\phi^2\\
				\end{bmatrix}
				$}
			& Observation uncertainty. Large uncertainty due to inherent measurement\newline noise. \textcolor{black}{{$\sigma_r \text{ = } 20\text{ cm, } \sigma_\phi^2 \text{ = } 0.5 \text{ rad$^2$}$}}\\
			\hline
			$\alpha_{t}$ \newline $\alpha_{r}$& \textcolor{black}{$=4$ \newline $=1$} & Mahalanobis distance thresholds.
			\newline $\alpha_{t}$ for filtering anomalous AOA observations (refer Section \ref{sect_ao}).
			\newline $\alpha_{r}$ for assigning the unknown correspondences (refer Section \ref{sect_sb}).
			
			\\
			\hline
		\end{tabular}
		\label{tab1}
	\end{center}
\end{table}

\subsection{All-UWB SLAM in a Feature-deficient Environment}

\subsubsection{\textcolor{black}{Experiment \#1 in a Vision-denied Environment}}
The TurtleBot was teleoperated at 5 cm/s in a U-shaped path, and the ground truth was obtained from the LiDAR\#1 scan data using the ROS2 SLAM toolbox (see Fig. \ref{ex_lab}). 
As a rule-of-thumb, a tag is deployed at the very beginning, regardless of the feature density in the environment. Consequently, this initial tag (i.e. artificial landmark) has a very low position uncertainty. Since the tags are detectable from afar unlike natural environmental features, the initial artificial landmark with low uncertainty enables the robot to reduce significant odometry drift during loop closures. More tags are later deployed in feature-deficient areas (refer Section \ref{sect_fda}).

Although the proposed All-UWB SLAM initially adheres to the odometry, it continues to rely on the tags and new features while refining the robot pose. The accuracy of the proposed systems was evaluated by finding the root mean squared (RMS) absolute trajectory error (ATE) after aligning the estimated poses with the ground truth from SLAM toolbox. The aligned poses and the generated landmark map is overlaid on the ground truth as shown in Fig. \ref{ex_lab}. Both ground truth and the All-UWB SLAM have started and stopped at the same location with a ATE of 10.3 cm.

\subsubsection{\textcolor{black}{Experiment \#2 in a Larger Environment}}
\textcolor{black}{In this experiment, the robot was teleoperated along a circular path in a larger feature-deficient environment. Similar to the previous experiment, the first tag was deployed at the very begining, and three more tags were deployed in distinct locations while exploring the environment. Finally, the pose estimations and the resulting landmark map were aligned and overlaid on the ground truth from the SLAM toolbox (see Fig. \ref{ex_hall}). Notice that both ground truth and the All-UWB SLAM pose estimations have started and stopped at the same location with a RMS ATE of only 7 cm.}

Moreover, the results clearly shows that the UWB radar has identified only the most prominent point features in the environment by neglecting the clutter. Those point features have been initialized as landmarks in the SLAM backend. Table \ref{tab1} summarizes the important hyper-parameters used in the experiment. It is important to note that these parameters are interrelated, and may require slight adjustments depending on the environment. As an example, gradual curves in the environment may occasionally get misidentified as point features. To avoid this, both window sizes: $m$ and $n$ must be increased. Subsequently, the clustering parameters must also be adjusted relative to $n$. Hence, a moderate understanding of the environment and intuition behind parameter tuning are required for optimal performance of All-UWB SLAM.

\subsection{\textcolor{black}{Ablation Studies}}
\textcolor{black}{
	Ablation studies were conducted to evaluate the codependency of UWB radar and UWB AOA in the proposed SLAM system, using the same \textit{rosbag} dataset from experiment \#1. 
}

\subsubsection{\textcolor{black}{Influence of UWB Radar}}
\textcolor{black}{
	This section omits the UWB radar observations and performs SLAM relying only on range and bearing observations from the anchor-tag units. As shown in Fig. \ref{ex_abs} (left), the robot was able to mitigate the odometry drift with respect to the first tag. However, in the middle region of the U-shaped environment, severe odometry drift caused the robot to completely lose its pose estimation accuracy. Consequently, the second tag was deployed and initialized in a wrong location. Hence, it is evident that relying solely on UWB AOA readings is insufficient while exploring an unknown environment.
}
\subsubsection{\textcolor{black}{Influence of UWB AOA Anchor-tags}}
\begin{figure}[!t]
	\centering
	{\includegraphics[clip,trim=0 0in 0 0, width=3.45in]{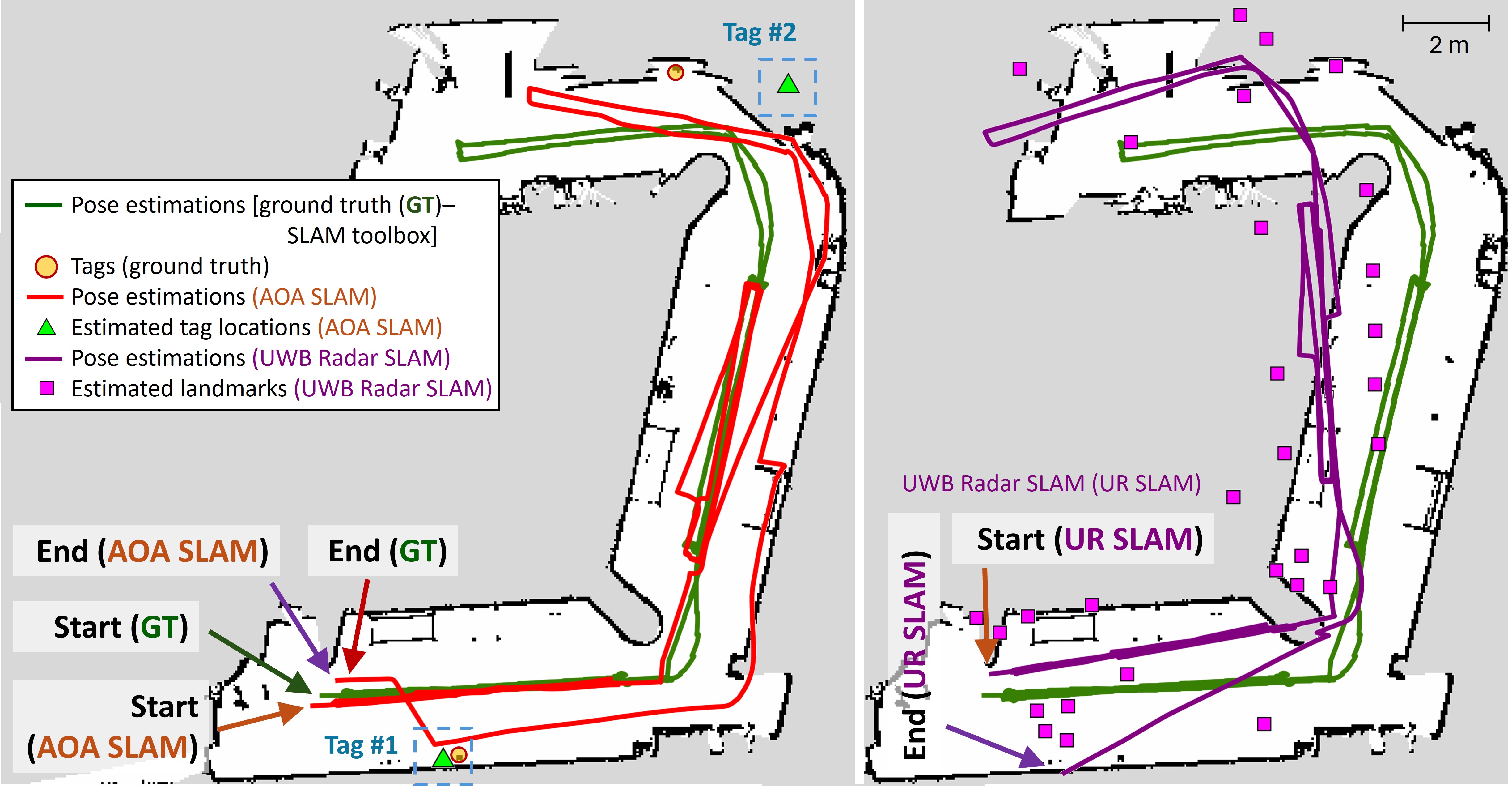}}
	\caption{\textcolor{black}{\textbf{Ablation studies.} The landmark map from the UWB AOA-only SLAM is aligned and overlaid on the occupancy grid map from the SLAM toolbox (left). Similarly, the landmark map from the vanilla UWB radar SLAM is overlaid on the SLAM toolbox map (right). Notably, the pose estimations in both cases show significant deviation from the ground truth.}
	}
	\label{ex_abs}
\end{figure}

\textcolor{black}{
	In this study, observations from UWB AOA anchor-tags were excluded, allowing the SLAM system to rely solely on UWB radar (i.e. vanilla UWB radar SLAM). The results in Fig. \ref{ex_abs} (right) demonstrate that UWB radar alone is incapable of performing correct loop closures in the test environment. Although the robot could perform SLAM in the feature-rich middle region of the environment, it was not able to recorrect its pose in close proximity to the starting point.} This erratic behavior can be attributed to two major reasons: 1) significant odometry drift, and 2) the correspondence of each natural landmark being unknown. On the other hand, UWB AOA anchor-tags are detectable from a long distance and have known correspondences. Hence, the robot can refine its pose error early before it becomes significant. 

\section{Conclusions} \label{sec::conclusions}

In this paper, we presented a novel approach for SLAM in vision-denied environments using UWB technology: All-UWB SLAM. The proposed system employs both UWB radar observations and UWB AOA anchor-tags to create a landmark map of an unknown feature-deficient environment in real-time. Artificial landmarks (i.e. UWB AOA nodes) are deployed in the identified feature-deficient areas to facilitate optimal SLAM performance.

\textcolor{black}{
	Experiments were performed in two unstructured indoor environments comprising only a few prominent point features.} The results demonstrate that the UWB AOA information improves UWB radar-based SLAM, enabling its application in more complex feature-deficient environments. The proposed system has successfully identified only the prominent point features using UWB radar, and eliminated UWB AOA ghost readings caused by NLOS conditions. In the future, we intend to integrate autonomous exploration using UWB radar for vision-denied environments,
	\textcolor{black}{and to propose a optimum sensor deployment scheme for UWB AOA nodes \cite{gdop}.}



%

\bibliographystyle{IEEEtran}
\bibliography{mybib}

\begin{thebibliography}{10}
\providecommand{\url}[1]{#1}
\csname url@samestyle\endcsname
\providecommand{\newblock}{\relax}
\providecommand{\bibinfo}[2]{#2}
\providecommand{\BIBentrySTDinterwordspacing}{\spaceskip=0pt\relax}
\providecommand{\BIBentryALTinterwordstretchfactor}{4}
\providecommand{\BIBentryALTinterwordspacing}{\spaceskip=\fontdimen2\font plus
\BIBentryALTinterwordstretchfactor\fontdimen3\font minus
  \fontdimen4\font\relax}
\providecommand{\BIBforeignlanguage}[2]{{%
\expandafter\ifx\csname l@#1\endcsname\relax
\typeout{** WARNING: IEEEtran.bst: No hyphenation pattern has been}%
\typeout{** loaded for the language `#1'. Using the pattern for}%
\typeout{** the default language instead.}%
\else
\language=\csname l@#1\endcsname
\fi
#2}}
\providecommand{\BIBdecl}{\relax}
\BIBdecl

\bibitem{RN378}
M.~Chghaf, S.~Rodriguez, and A.~E. Ouardi, ``Camera, lidar and multi-modal slam
  systems for autonomous ground vehicles: a survey,'' \emph{Journal of
  Intelligent \& Robotic Systems}, vol. 105, no.~1, p.~2, 2022.

\bibitem{10320444}
Z.~Jian \emph{et~al.}, ``Path generation for wheeled robots autonomous
  navigation on vegetated terrain,'' \emph{IEEE Robotics and Automation
  Letters}, vol.~9, no.~2, pp. 1764--1771, 2024.

\bibitem{RN339}
Z.~Hong, Y.~Petillot, A.~Wallace, and S.~Wang, ``Radarslam: A robust
  simultaneous localization and mapping system for all weather conditions,''
  \emph{The International Journal of Robotics Research}, vol.~41, no.~5, pp.
  519--542, 2022.

\bibitem{smoke}
W.~Chen \emph{et~al.}, ``Constructing floor plan through smoke using ultra
  wideband radar,'' \emph{Proc. ACM Interact. Mob. Wearable Ubiquitous
  Technol.}, vol.~5, no.~4, Dec. 2022.

\bibitem{cp1}
H.~A. G.~C. Premachandra, R.~Liu, C.~Yuen, and U.~X. Tan, ``Uwb radar slam: an
  anchorless approach in vision denied indoor environments,'' \emph{IEEE
  Robotics and Automation Letters}, pp. 1--8, 2023.

\bibitem{liu2024rangeslamultrawidebandbasedsmokeresistantrealtime}
\BIBentryALTinterwordspacing
Y.~Liu,  \emph{et~al.}, ``Range-slam: Ultra-wideband-based smoke-resistant
  real-time localization and mapping,'' 2024. [Online]. Available:
  \url{https://arxiv.org/abs/2409.09763}
\BIBentrySTDinterwordspacing

\bibitem{RN83}
D.~Wang, S.~Yoo, and S.~H. Cho, ``Experimental comparison of ir-uwb radar and
  fmcw radar for vital signs,'' \emph{Sensors}, vol.~20, no.~22, p. 6695, 2020.

\bibitem{RN60}
G.~Schouten and J.~Steckel, ``Radarslam: Biomimetic slam using ultra-wideband
  pulse-echo radar,'' in \emph{2017 International Conference on Indoor
  Positioning and Indoor Navigation (IPIN)}, pp. 1--8.

\bibitem{RN123}
T.~Deißler and J.~Thielecke, ``Uwb slam with rao-blackwellized monte carlo
  data association,'' in \emph{2010 International Conference on Indoor
  Positioning and Indoor Navigation}, pp. 1--5.

\bibitem{RN139}
------, ``Fusing odometry and sparse uwb radar measurements for indoor slam,''
  in \emph{2013 Workshop on Sensor Data Fusion: Trends, Solutions, Applications
  (SDF)}, pp. 1--5.

\bibitem{radio}
Amjad \emph{et~al.}, ``Radio slam: A review on radio-based simultaneous
  localization and mapping,'' \emph{IEEE Access}, vol.~11, pp. 9260--9278,
  2023.

\bibitem{feng}
F.~Ge and Y.~Shen, ``Single-anchor ultra-wideband localization system using
  wrapped pdoa,'' \emph{IEEE Transactions on Mobile Computing}, vol.~21,
  no.~12, pp. 4609--4623, 2022.

\bibitem{zhou}
H.~Zhou, Z.~Yao, and M.~Lu, ``Lidar/uwb fusion based slam with
  anti-degeneration capability,'' \emph{IEEE Transactions on Vehicular
  Technology}, vol.~70, no.~1, pp. 820--830, 2021.

\bibitem{RN58}
J.~Tiemann, A.~Ramsey, and C.~Wietfeld, ``Enhanced uav indoor navigation
  through slam-augmented uwb localization,'' in \emph{2018 IEEE International
  Conference on Communications Workshops}, pp. 1--6.

\bibitem{RN379}
Y.~Zhang, J.~Li, and S.~Zhao, ``Fusion positioning of ultra-wideband single
  base station and visual inertial odometry based on arrival of angle and time
  of flight,'' in \emph{2023 2nd International Conference on Robotics,
  Artificial Intelligence and Intelligent Control (RAIIC)}, 2023, pp. 365--371.

\bibitem{fire}
S.~Wang, P.~Xu, W.~Weng, L.~Niu, and R.~Wang, ``An end-to-end recognition
  method for ir-uwb radar dynamic detection mode for detecting targets in fire
  rescue scenarios,'' \emph{IEEE Internet of Things Journal}, vol.~11, no.~23,
  pp. 38\,137--38\,150, 2024.

\bibitem{wang}
T.~Wang, H.~Zhao, and Y.~Shen, ``An efficient single-anchor localization method
  using ultra-wide bandwidth systems,'' \emph{Applied Sciences}, vol.~10,
  no.~1, 2020.

\bibitem{aoa_1}
C.-M. Zhong \emph{et~al.}, ``Uwb-based aoa indoor position-tracking system and
  data processing algorithm,'' \emph{IEEE Sensors Journal}, vol.~24, no.~19,
  pp. 30\,522--30\,529, 2024.

\bibitem{ROSLAM}
A.~Torres-González, J.~R. Martinez-de Dios, and A.~Ollero, ``Range-only slam
  for robot-sensor network cooperation,'' \emph{Autonomous Robots}, vol.~42,
  no.~3, pp. 649--663, 2018.

\bibitem{uwblidar}
Y.~Song, M.~Guan, W.~P. Tay, C.~L. Law, and C.~Wen, ``Uwb/lidar fusion for
  cooperative range-only slam,'' in \emph{2019 International Conference on
  Robotics and Automation (ICRA)}, 2019, pp. 6568--6574.

\bibitem{sensorfly}
X.~Chen \emph{et~al.}, ``Design experiences in minimalistic flying sensor node
  platform through sensorfly,'' \emph{ACM Trans. Sen. Netw.}, vol.~13, no.~4,
  Nov. 2017.

\bibitem{drunkwalk}
------, ``H-drunkwalk: Collaborative and adaptive navigation for heterogeneous
  mav swarm,'' \emph{ACM Trans. Sen. Netw.}, vol.~16, no.~2, Apr. 2020.

\bibitem{aoa}
\BIBentryALTinterwordspacing
\emph{LinkTrack AOA Datasheet}, Nooploop, 2020, version 1.1. [Online].
  Available:
  \url{https://ftp.nooploop.com/downloads/linktrack_aoa/LinkTrack_AOA_Datasheet_V1.1_zh.pdf}
\BIBentrySTDinterwordspacing

\bibitem{angu}
M.~Heydariaan, H.~Dabirian, and O.~Gnawali, ``Anguloc: Concurrent angle of
  arrival estimation for indoor localization with uwb radios,'' in \emph{2020
  16th International Conference on Distributed Computing in Sensor Systems
  (DCOSS)}, Conference Proceedings, pp. 112--119.

\bibitem{gdop}
L.~Santoro, D.~Brunelli, and D.~Fontanelli, ``On-line optimal ranging sensor
  deployment for robotic exploration,'' \emph{IEEE Sensors Journal}, vol.~22,
  no.~6, pp. 5417--5426, 2022.

\end{thebibliography}

%




\end{document}